\documentclass{article}

\usepackage[nonatbib,preprint]{neurips_2020}

\usepackage[utf8]{inputenc} % allow utf-8 input
\usepackage[T1]{fontenc}    % use 8-bit T1 fonts
\usepackage{hyperref}       % hyperlinks
\usepackage{url}            % simple URL typesetting
\usepackage{booktabs}       % professional-quality tables
\usepackage{amsfonts}       % blackboard math symbols
\usepackage{nicefrac}       % compact symbols for 1/2, etc.
\usepackage{microtype}      % microtypography

\usepackage{amsmath,amsthm}
\usepackage{algorithm}% http://ctan.org/pkg/algorithms
\usepackage{algorithmic}% http://ctan.org/pkg/algorithmicx
\usepackage{amssymb} 
\usepackage{graphicx}
\usepackage{subfigure}
\usepackage{blindtext}
\usepackage{multirow}
\usepackage{paralist}
\usepackage{wrapfig,lipsum,booktabs}
\usepackage{soul}
\usepackage[titletoc]{appendix}

\usepackage{xcolor}

\newcommand{\veca}{\ensuremath{\mathbf{a}}}
\newcommand{\vecb}{\ensuremath{\mathbf{b}}}
\newcommand{\x}{\ensuremath{\mathbf{x}}}
\newcommand{\z}{\ensuremath{\mathbf{z}}}
\newcommand{\h}{\ensuremath{\mathbf{h}}}
\newcommand{\f}{\ensuremath{\mathbf{f}}}

\newcommand{\actions}{\ensuremath{\mathcal{A}}}

\newtheorem{definition}{Definition}

\graphicspath{{../}}

\title{Reinforcement Learning with Efficient \\
Active Feature Acquisition}

\author{%
  Haiyan Yin\thanks{Work done in part while at Microsoft Research Cambridge.}\\
  Nanyang Technological University\\
  Singapore
  \And
  Yingzhen Li\\
  Microsoft Research\\
  United Kingdom
  \And
  Sinno Jialin Pan\\
  Nanyang Technological University\\
  Singapore
 \And
  Cheng Zhang\\
  Microsoft Research\\
  United Kingdom
  \And
  Sebastian Tschiatschek$^*$\\
  University of Vienna\\
  Austria  
}

\begin{document}

\maketitle

\begin{abstract}
Solving real-life sequential decision making problems under partial observability involves an exploration-exploitation problem. To be successful, an agent needs to efficiently gather valuable information about the state of the world for making rewarding decisions. However, in real-life, acquiring valuable information is often highly costly, e.g., in the medical domain, information acquisition might correspond to performing a medical test on a patient. This poses a significant challenge for the agent to perform optimally for the task while reducing the cost for information acquisition. 
In this paper, we propose a model-based reinforcement learning framework that learns an active feature acquisition policy to solve the exploration-exploitation problem during its execution. Key to the success is a novel sequential variational auto-encoder that learns high-quality representations from partially observed states, which are then used by the policy to maximize the task reward in a cost efficient manner. We demonstrate the efficacy of our proposed framework in a control domain as well as using a medical simulator. 
In both tasks, our proposed method outperforms conventional baselines and results in policies with greater cost efficiency.
\end{abstract}

\section{Introduction}
\label{sec:intro}

Recently, machine learning models for automated sequential decision making have shown remarkable success across many application areas, such as visual recognition~\cite{das2017learning,mathe2016reinforcement}, robotics control~\cite{finn2016guided,zhang2018solar}, medical diagnosis~\cite{ling2017learning,peng2018refuel} and computer games~\cite{mnih-dqn-2015,silver2016mastering}.
%One fundamental reason that drives the success of such models and enables them to outperform classical algorithms is the availability of large amounts of training data.
%Typically this training data is either fully observed or the features stem from an action-independent observation model (which clearly can depend on the state of the system).
These models are typically trained on large amounts of data with a fixed set of available features, and when these models are deployed, they are assumed to operate on data with the same features.
However, in many real-world applications, the fundamental assumption that the same features are always readily available during deployment does not hold or it is desired that the model can also operate on different sets of features.
For instance, consider a medical support system for monitoring and treating patients during their stay at hospital,
% \CZ{maybe give an example with time such as patient staying in hospital} 
which was trained on rich historical medical data.
%that needs to make a diagnosis for a patient \CZ{instead of diagnosis maybe monitoring the health situation at any time. Just make this example more similar to the sepsis simulator one and have the time component}.
To provide the best possible treatment, the system might need to perform several measurements of the patient over time. 
However, some of these measurements could be costly or pose a health risk.
That is, during deployment, the system should function with minimal and carefully selected features while during training more features might have been available.
% Thus, it is impractical to train the decision making model over the set of all the available measurements.
Hence, under such scenarios, we are interested in decision making models that take the measurement process, i.e., feature acquisition, into account and only acquire the information relevant for making a treatment decision.

In this paper, we consider the challenging problem of learning effective sequential decision making policies when the cost of information acquisition cannot be neglected. 
% To be successful, we need to learn policies which acquire the information required for solving a task in the most cost efficient way.
To be successful, we need to learn policies which acquire the information required for making the task related decisions in the most cost efficient way.
For simplicity, we can think of the policies as being constituted of an \emph{acquisition policy}, which selects the features to be observed and a \emph{task policy}, which selects actions to change the state of the system towards some goal. As a consequence, these two policies are typically intimately connected, i.e., the acquisition policy must collect features such that the task policy can take good actions, and the task policy needs to enable the acquisition policy to collect informative features by transiting to appropriate states.
As such, our work tackles a partially observable policy learning problem with the following two distinguishing properties compared to the most commonly studied problems.
% (see also Figure~\ref{fig:pomdp} for an illustration).
%\begin{inparaenum}[\itshape (i)\upshape]
\vspace{-6pt}
\begin{itemize}[\itshape (i)\upshape]
\item[1.] By incorporating active feature acquisition, the training of the task policy is based upon subsets of features only, i.e., there are missing features, where the missingness is controlled by the acquisition policy. 
Thus, the resulting POMDP is different from typically considered POMDPs in RL literature~\cite{cassandra1998survey} where the partial observability stems from a fixed and action-independent observation model. 
Also, the state-transitions in conventional POMDPs are often only determined by the choice of the task action, whereas in our setting the state-transition is affected by both the task action and the feature acquisition choice.
\item[2.] The learning of the acquisition policy introduces an additional dimension to the exploration-exploitation problem: each execution of the acquisition and task policy needs to solve an exploration-exploitation problem, and thus we often need to learn sophisticated policies.
%to perform well during evaluation, the algorithm needs to search from both the space of task actions as well as the feature acquisition actions, and thus this becomes more challenging.
\end{itemize}
\vspace{-6pt}
%\end{inparaenum}

Most reinforcement learning research has not taken active feature acquisition into consideration. In this work, we propose a unified approach that jointly learns a policy for optimizing the task reward while performing active feature acquisition.
Although some of the prior works have exploited the use of reinforcement learning for sequential feature acquisition tasks~\cite{shim2018joint,zannone2019odin}, they considered variable-wise information acquisition in a static setting only, corresponding to feature selection for non-time-dependent prediction tasks.
%Formulating sequential acquisition over such problems would result in incremental observed variables, with a fixed statistic base.
% However, in our problem, we consider the more challenging sequential decision making over time-series data, where an acquisition is made at each time step along the trajectory to select a feature subset. As such, both the model dynamics of the underlying MDP and the choice of feature acquisition introduced considerable challenges to learning the sequential feature acquisition strategy. 
However, our considered setting is truly time-dependent and the decision of feature acquisition needs to be made at each time step while the state of the system evolves simultaneously.
% As such, both the model dynamics of the underlying MDP and the choice of feature acquisition introduce considerable challenges to learning the sequential feature acquisition strategy.

Due to the challenge of the exploration-exploitation problem, it is a non-trivial task to jointly learn the policies. 
% The conventional end-to-end approaches often result in inferior solutions in complex scenarios. Ideally, policies based on high-quality representations would be easier for the algorithm to search for better solutions through exploration-exploitation. 
We approach this problem and present a framework which tackles the problem from a representation learning perspective. In particular, we make the following contributions:
1. We propose a general solution for learning reinforcement learning policies with active feature acquisition. Our proposed approach simultaneously learns reinforcement learning policies for reward optimization and active feature acquisition, approximately solving a challenging combinatorial problem. 2. We present a novel sequential representation learning approach to account for the encoding of the partially observed states based on sequential variational autoencoders (VAE). 3. We present experiment results on an image-based control task as well as a medical simulator fitted from real-life data, where our method shows clear improvements over natural baselines.

%% Related Work
%++++++++++++++++++++++++++++++++++++++++++++++++++
\section{Related Work}
\label{sec:related_work}

Our work jointly considers active learning and reinforcement learning, to accomplish the policy training task while acquiring fewer observed features as possible. We thus review related methods for active feature acquisition and representation learning for POMDP, respectively.

\vspace{-5pt}
\paragraph{Active Feature Acquisition} 
% Our work draws motivation from the existing instance-wise active feature selection approaches. One category of the instance-wise feature selection methods consider feature acquisition as a one time effort to select a subset of features at each time. A typical example is the conventional linear model that poses sparsity inducing prior distribution to the model~\cite{tibshirani1996regression}. Recently, there also emerged approaches that adopt reinforcement learning to actively find optimal feature subsets~\cite{shim2018joint, yoon2018invase, zannone2019odin}. Though such attempts have demonstrated certain efficacy in handling non time-series instance-wise data, they do not suffice for handling sequential dataset. There is an alternative category that models feature acquisition as a Bayesian experiment design~\cite{GongTNTHZ19,ma2020vaem, MaTPHNZ19}. However, the sequential decision making is for variable-wise feature acquisition and the problems are still non time-series tasks in nature. The key difference between all the above approaches and ours is in that we tackle active feature acquisition for the time-series tasks, where an active feature selection decision needs to be formed at each time step along the multi-step reinforcement learning trajectory. Therefore, the feature acquisition for our presented work needs to consider more complex information over model dynamics and control, apart from the static instance-wise features.  

Our work draws motivation from the existing instance-wise active feature selection approaches. One category of the instance-wise feature selection methods consider feature acquisition as a one time effort to select a subset of features at each time. One typical example is the conventional linear model that poses sparsity inducing prior distribution to the model~\cite{tibshirani1996regression}. There are also a number of approaches that adopt reinforcement learning to actively find optimal feature subsets, with successful applications in the fields of active perception/sensor selection~
\cite{satsangi2020maximizing,spaan2009decision}, visual object localization/tracking~\cite{jie2016tree,yun2018action}, medical diagnosis~\cite{yoon2018invase, zannone2019odin} and many more. While the aforementioned works focus on selecting a feature subset at a time, there is an alternative category that models feature acquisition as a sequential Bayesian experiment design~\cite{GongTNTHZ19,ma2020vaem, MaTPHNZ19}. However, the sequential decision making specified in those works performs variable-wise feature acquisition in a sequential manner, so the problems are still non time-series tasks in nature. 
In this paper, our work tackles active feature acquisition for time-series tasks. Moreover, unlike most of the existing reinforcement learning-based active feature acquisition works which learn a policy for active feature acquisition only, our work considers simultaneously learning a reinforcement learning policy and an active feature acquisition policy. 
%The problem we consider is settled on a more complicated system dynamics than the aforementioned works, as performing feature acquisition would greatly reduce the degree of observability for agent when learning task skills. 

\vspace{-5pt}
\paragraph{Representation Learning in POMDP}
The task of simultaneously learning reinforcement learning policy and active feature acquisition policy would result in a policy training scenario with partial observability, for which learning meaningful representation would become an essential and non-trivial research challenge. 
% When considering problems with POMDPs, the state space is partially presented to the agent, which makes representation learning an essential and non-trivial research challenge. 
Most conventional (deep) reinforcement learning approaches unifies the process of representation learning with policy training and results in policies trained in an end-to-end fashion~\cite{lillicrap2015continuous,mnih2016asynchronous,mnih2013playing}. However, such models often engage trainable parameters with considerable size and result to be less sample efficient. Another prominent line of research tackles the representation learning for POMDP in an off-line fashion, which results in multi-stage reinforcement learning.  In~\cite{higgins2016beta,higgins2017darla}, pretrained VAE models are adopted as the representation module to build agents with strong domain adaptation performance. The key difference between their works and ours is in that they consider typical POMDP domains where the state presents partial view over the environment, whereas ours considers a unique setting of partial observability, i.e., feature-level information could be \emph{missing} at each step. We thus adopt a sequential representation learning approach to infer a more informative state representation. Recently, there emerged a fruitful literature over sequential representation learning for POMDP~\cite{gregor2018temporal,vezzani2019learning}, where most of them formulate VAE training as an auxiliary task to optimize the representation model jointly with the policy training. In our work, we consider representation learning for partially observed sequences with a model-based attempt, where a novel sequential generative model is trained to learn model dynamics and generate high-quality feature representations. 
% We demonstrate such representation learning method plays a vital role in generating high-quality features to facilitate the joint policy training task.
Our attempt of learning model dynamics to gather information over the unobserved features is also related to image inpainting works~\cite{MatteiF19, yeh2017semantic, zheng2019disentangling}. However, such methods mostly focus on inpainting static images, such as face images, whereas we consider imputing the features from time-series data. Another primary difference between our work and theirs is that we focus on simultaneously learning reinforcement learning and active feature acquisition policies, rather than considering image inpainting only.

%% Methodology
% \input{methodology.tex}
\section{Methodology}
\label{sec:method}
%++++++++++++++++++++++++++++++++++++++++++++++++++

%++++++++++++++++++++++++++++++++++++++++++++++++++

\subsection{Problem Setting}
In this section, we formalize our problem setting. To this end, we define the \emph{active feature acquisition POMDP}, a rich class of discrete-time stochastic control processes generalizing standard POMDPs:
\begin{definition}[AFA-POMDP]
  The \emph{active feature acquisition POMDP} is a tuple $\mathcal{M} = \langle \mathcal{S}, \mathcal{A}, \mathcal{T}, \mathcal{O}, \mathcal{R}, \mathcal{C}, \gamma \rangle$, where $\mathcal{S}$ is the state space and $\mathcal{A}=\mathcal{A}^c \times \mathcal{A}^f$ is a joint action space of  feature acquisition actions $\mathcal{A}^f$ and control actions $\mathcal{A}^c$. 
  The transition kernel $\mathcal{T}\colon \mathcal{S} \times \mathcal{A}^c \times \mathcal{A}^f \rightarrow P_\mathcal{S}$ maps any joint action $\veca=(\veca^c, \veca^f)$ in state $s \in \mathcal{S}$ to a distribution $P_\mathcal{S}$ over next states.
  In each state $s$, the agent observes the features $\x^p$ which are a subset of the features $\x = (\x^p, \x^u) \sim \mathcal{O}(s)$ selected by the agent taking feature acquisition action $\mathbf{a}^f$, where $\mathcal{O}(s)$ is a distribution over possible feature observation for state $s$ and $\x^u$ are features not observed by the agent.
  When taking a joint action, the agent obtains rewards according to the reward function $\mathcal{R}\colon \mathcal{S} \times \mathcal{A}^c \rightarrow \mathbb{R}$ and pays a cost of $\mathcal{C} \colon \mathcal{S} \times \mathcal{A}^f \rightarrow \mathbb{R}_{\geq 0}$ for feature acquisition.
  Rewards and costs are discounted by the discount factor $\gamma \in [0,1)$.
\end{definition}
% By incorporating the active feature acquisition process, the policy training problem naturally becomes a particular partially observable Markov decision process (POMDP), with the task policy being trained only on an actively selected subset of features. 

\paragraph{Simplifying assumptions} 
For simplicity, we assume that $\x$ consists of a fixed number of features $N_f$ for all states, that $\mathcal{A}^f = 2^{[N_f]}$ is the powerset of all the $N_f$ features, and that $\x^p(\mathbf{a}^f)$ consists of all the features in $\x$ indicated by the subset $\mathbf{a}^f \in \mathcal{A}^f$. 
% The control action space is $\actions^c \subseteq \mathbb{R}^{|A_c|}$.
Note that the feature acquisition action for a specific application can take various different forms.
For instance, in our experiments in Section~\ref{sec:exp}, for the \emph{Sepsis} task, we define feature acquisition as selecting a subset over possible \emph{measurement} tests, whereas for the \emph{Bouncing Ball}$^+$ task, we divide an image into four observation regions and let the feature acquisition policy select a subset of observation regions (rather than raw pixels).
Please also note that while in a general AFA-POMDP, the transition between two states depends on the joint action, we assume in the following that it depends only on the control action, i.e., $\mathcal{T}(s, \veca^c, \veca^{f'}) = \mathcal{T}(s, \veca^c, \veca^{f})$ for all $\veca^{f'},\veca^{f} \in \mathcal{A}^f$. 
While not true for all possible applications, this assumption can be a reasonable approximation for instance for medical settings in which tests are non-invasive. 
For simplicity we furthermore assume that acquiring each feature has the same cost, denoted as $c$, i.e., $\mathcal{C}(\mathbf{a}^f, s)= c \, |\mathbf{a}^f| $, but our approach can be straightforwardly adapted to have different costs for different feature acquisitions.

\paragraph{Objective}
We aim to learn a policy which trades off reward maximization and the cost for feature acquisition by jointly optimizing a task policy $\pi^c$ and a feature acquisition policy $\pi^f$.
%The optimization of the task reward is grounded by the reward function $\mathcal{R}(s, \veca^c) \in \mathbb{R}$.
%, which yields a real-valued reward for the pair of true MDP state $s$ and task action $\veca^c$.
%The feature acquisition cost is r.
That is, we aim to solve the optimization problem
\begin{equation}
\label{eq:objective}
\max_{\pi^f, \pi^c} \quad \mathbb{E} \!\! \left[ \sum_{t=0}^\infty \gamma^t \Big( \mathcal{R}(s_t, \veca^c_t) - \sum_i^{|\actions_f|} c\cdot \mathbb{I}\,(\veca^{f(i)}_t)\Big)  \right] \!\!,
\end{equation}
where the expectation is over the randomness of the stochastic process and the policies, $s_t$ is the state of the system at time $r$, $\veca^{f(i)}_t$ denotes the $i$-th feature acquisition action at time $t$, and $\mathbb{I}\,(\cdot)$ is an indicator function whose value equals to 1 if that feature has been acquired. 
% \sebastian{The $f(i)$ notation is not properly introduced.}

The above optimization problem is very challenging: an optimal solution needs to maintain beliefs $\vecb_t$ over the state of the system at time $t$ which is a function of partial observations obtained so far. Both the the feature acquisition policy $\pi^f(\veca^f_t \mid \vecb_t)$ and the task policy $\pi^c(\veca^c_t \mid \vecb_t)$ depend on this belief.
%The task policy $\pi^c$ can only condition on the observed partial features so far, which possibly conveys only very limited information. 
%$\pi^f$ is trained to reduce the cost at a significant scale.
The information in the belief itself can be controlled by the feature acquisition policy through querying subsets from the features $\x_t$ and hence the task policy and feature acquisition policy itself strongly depend on the effectiveness of the feature acquisition.
Through enabling to query subsets of observations, the feature acquisition action space $\mathcal{A}^f$ is exponential in the number of features.

\paragraph{Remarks}
Clearly, any AFA-POMDP corresponds to a POMDP in which the reward is defined appropriately from $\mathcal{R}$ and $\mathcal{C}$ and observations depend on the taken joint action. 
In principle this provides a natural way for approaching AFA-POMDPs: map them to the corresponding POMDP and (approximately) solve this POMDP using any suitable method.
There is however an additional challenge because of the exponential size of the feature acquisition state space.
In many practical applications this explosion, however, is not that severe.
For instance in many medical applications, there are only a few costly or dangerous measurements while other information like demographics or a person's temperature are available at essentially no cost.
General scaling of RL to large action spaces is an interesting and active research topic orthogonal to our work.
Studying hierarchical representations of the measurements for feature selection in the context of AFA-POMDPs, which can likely alleviate issues due to the large action space, are subject to future work.
% This can often be achieved by performing a suitable diagnostic test taken from a small subset of all possible diagnostic tests, thereby drastically reducing the otherwise exponential action space. 

%++++++++++++++++++++++++++++++++++++++++++++++++++
% \vspace{-2mm}
\subsection{Sequential Representation Learning with Partial Observations}

\begin{wrapfigure}{r}{0.5\columnwidth}
\centering
\vspace{-15pt}
\vspace{1mm}
\includegraphics[width=0.5\textwidth]{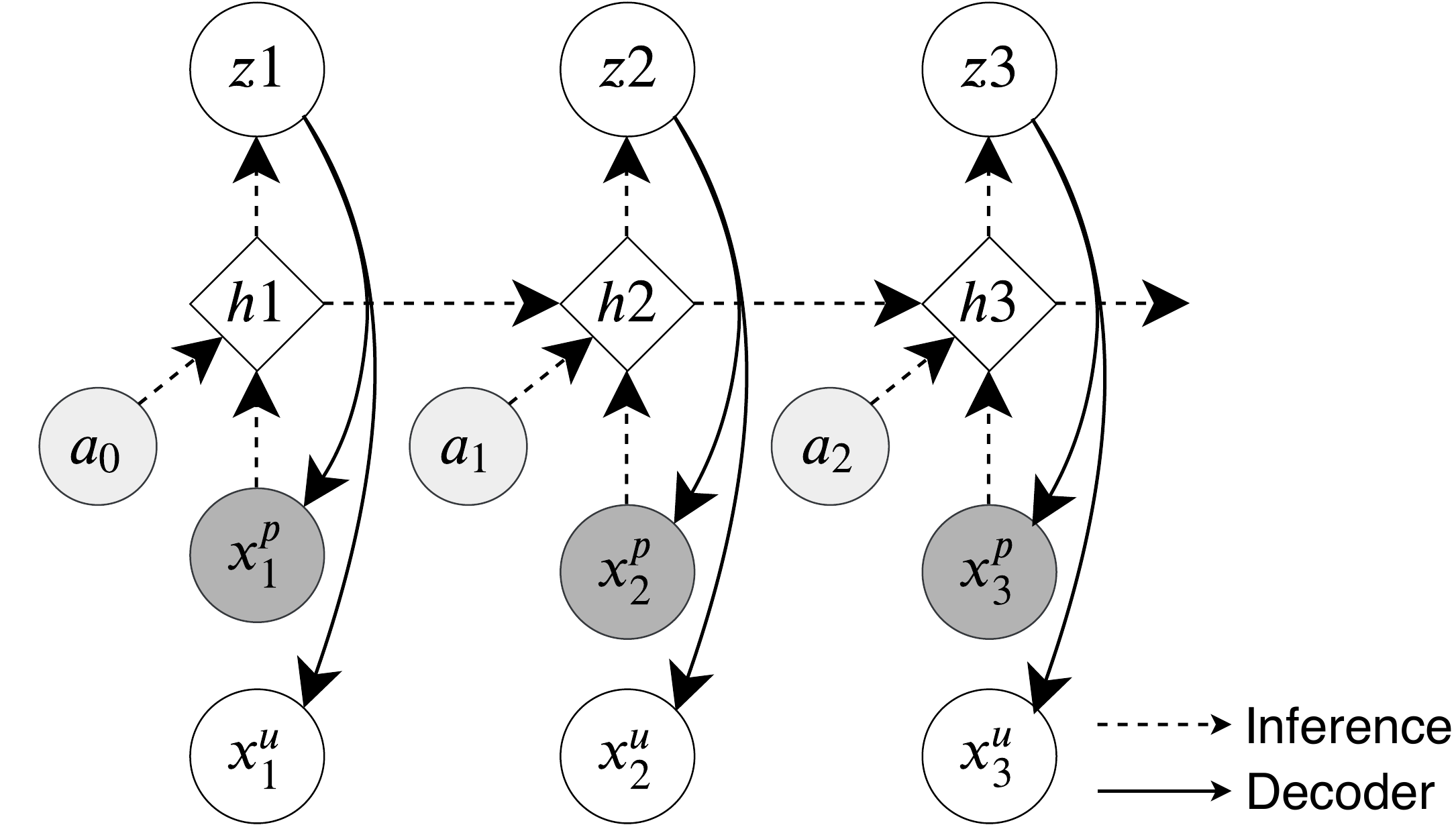}
\vspace{-15pt}
 \caption{
The model architecture for our proposed partially observable sequential VAE. Shaded nodes represent the observed variables. 
%   The inference model filters information over the partial observations and actions, to predict both the observed and unobserved features.
} \label{fig:vae}
\end{wrapfigure} 

We introduce a sequential representation learning approach to facilitate the task of policy training with active feature acquisition.
Let $\x_{1:T} = (\x_1, ..., \x_T)$ and $\veca_{1:T} = (\veca_1, ..., \veca_T)$ denote a sequence of observations and actions, respectively. Alternatively, we also denote these sequences as $\x_{\leq T}$ and $\veca_{\leq T}$. 
Overall, our task of interest is to train a sequential representation learning model to learn the distribution of the full sequential observations $\x_{1:T}$, i.e., for both the observed part $\x^p_{1:T}$ and the unobserved part $\x^u_{1:T}$. Given  only partial observations, we can perform inference only with the observed features $\x^p_{1:T}$. Therefore, our proposed approach extends the conventional unsupervised representation learning task to a supervised learning task, which learns to impute the unobserved features by synthesizing the acquired information from the history and learning the model dynamics.

Our key assumption is that learning to impute the unobserved features leads to better representations which can be leveraged by the task policy and that, because of partial observability, sequential representation learning is better than non-sequential learning. 
% Furthermore, unlike many conventional sequential representation learning models for reinforcement learning that only reason over the observation sequence $\x^p_{1:T}$, in our work, we take into account both the observation sequence $\x^p_{1:T}$ and the action sequence $\veca_{1:T}$ for conducting inference. The intuition is that since $\x^p_{1:T}$ by itself consists of very limited information over the agent's underlying MDP state, incorporating the action sequence would be an informative add-on to the agent's acquired information to infer the belief state. 
To summarize, our proposed sequential representation model learns to encode $\x^p_{1:T}$ and $\veca_{1:T}$ into meaningful latent features, for predicting $\x^p_{1:T}$ and $\x^u_{1:T}$. The architecture of our proposed sequential representation learning model is shown in Figure~\ref{fig:vae}.

% \begin{figure}[t!]
% 	\centering
% 	\begin{minipage}[c]{0.5\textwidth}
%     \includegraphics[width=0.9\textwidth]{latex/figures/ssm.pdf}
%   \end{minipage}\hfill
%   \begin{minipage}[c]{0.48\textwidth}
%     \caption{
%       Observation decoder and belief inference model for the partially observable sequential VAE. Shaded nodes represent the observed variables. The inference model filters information over the partial observations and actions, to predict both the observed and unobserved features.
%     } \label{fig:vae}
%   \end{minipage}
% \end{figure}

\paragraph{Observation Decoder} Let $\z_{1:T} = (\z_1, ..., \z_T)$ denote a sequence of latent states. We consider the following probabilistic model: %\sebastian{Is this missing the transition model? Or is there no transition model because of the arguments below?}
%\HY{There is no transition model being considered. The generative model itself is a standard VAE decoder without sequential modeling. }
\begin{equation}\label{eq:vae_gen}
    p_\theta(\x^p_{1:T}, \x^u_{1:T}, \z_{1:T}) = \prod_{t=1}^T p(\x_t^p, \x_t^u| \z_t) \, p(\z_t),
\end{equation}
% \sebastian{Mention/explain that there is no transition model.}
For simplicity of the notations, we assume $\z_0 = \textbf{0}$. We impose a simple prior distribution over $\z$, i.e., a standard Gaussian prior, instead of incorporating some learned prior distribution over the latent space of $\z$, such as an autoregressive prior distribution like $p(\z_t|\z_{t-1}, \x^p_{1:t}, \veca_{0:t-1})$. The reason is that using a static prior distribution results in latent representation $\z_t$ that is stronger regularized and more normalized than using a learned prior distribution which is stochastically changing over time. This is crucial for deriving stable policy training performance. 
% We claim that using a static prior distribution for $\z_t$ is crucial for deriving a stable policy learning performance when using the latent representation $\z_t$ later at the policy training phase. 
% \YL{I am confused, in the LHS of Figure 2, it shows a state-space model where $\z_t$ depends on $\z_{t-1}$, but in text you said $\z_t$ are independent?}
At time $t$, the generation of data $\x_t^p$ and $\x_t^u$ depends on the corresponding latent variable $\z_t$. Given $\z_t$, the observed variables are conditionally independent of the unobserved ones. Therefore, 
\begin{equation}
p(\x_t^p, \, \x_t^u|\z_t) = p(\x_t^p|\z_t) \, p(\x_t^u|\z_t).
\end{equation}

\paragraph{Belief Inference Model} 
During policy training, we only assume access to partially observed data. This requires an inference model which takes in the past observation and action sequences to infer the latent states $\textbf{z}$. Specifically, 
 we present a structured inference network $q_\phi$ as shown in Figure~\ref{fig:vae}, which has an autoregressive structure:
\begin{equation}
q_\phi(\z_{1:T}|\x_{1:T},\veca_{<T}) = \prod_{t=1}^T q_\phi(\z_t|\x_{\leq t}^p, \veca_{< t}),
\end{equation}
where $q_\phi(\cdot)$ is a function that aggregates the filtering posteriors of the history of observation and action sequences.
%\sebastian{@Haiyan: I changed the conditions in the previous equations to lower case t from T. Is that what you implemented?}\HY{Yes it should be t}
% \sebastian{This is a bit odd, as the generative model does not have temporal dependencies. I understand that we need the inference model in the form as we need to compute the belief. Let's just discuss this briefly tomorrow.}
Following the common practice in existing sequential VAE literature, we adopt a forward RNN model as the backbone for the filtering function $q_\phi(\cdot)$~\cite{gregor2018temporal}. Specifically, at step $t$, the RNN processes the encoded partial observation $\x_t^p$, action $\veca_{t-1}$ and its past hidden state $\h_{t-1}$ to update its hidden state $\h_t$. Then the latent distribution $\z_t$ is inferred from $\h_t$. The belief state $\vecb_t$ is defined as the mean of the distribution $\z_t$. By accomplishing the supervised learning task, the belief state could provide abundant information for not only the observed sequential features, but also for the missing features, so that the policy trained over it could benefit from it and progress faster towards getting better convergent performance. 

\paragraph{Learning} 
We proposed to pre-train both the generative and inference models offline before learning the RL policies. In this case, we assume the access to the unobserved features, so that we can construct a supervised learning task to learn to impute unobserved features. Concretely, the pre-training task update the parameters $\theta, \phi$ by maximizing the following variational lower-bound \cite{jordan1999introduction, kingma2013auto, zhang2018advances}:
\begin{align}
    \mbox{log}\,p(\x^p_{1:T},\x^u_{1:T})  \geq \; & \mathbb{E}_{q_\phi} \Big[ \sum_t \mbox{log} \, p_\theta(\x^p_t|\z_t) + \mbox{log} \, p_\theta(\x^u_t|\z_t) \nonumber 
    - \mbox{KL}\big(q_\phi(\z_t|\x^p_{\leq t}, \veca_{< t})\,||\,p(\z_t)\big) \Big] \nonumber\\
    = \; & \mbox{ELBO}(\x^p_{1:T},\x^u_{1:T}). \label{eq:elbo}
    %\textcolor{blue}{p_\theta(z_t|z_{t-1}, x_{\leq t}^p, a_{< t})}
\end{align}
%\sebastian{In the above equation, should we put $\x^p_{1:T}$ and $\x^u_{1:T}$?}\HY{Yes. Same for Eq(2)(4). Both have been updated.}
By incorporating the term $\mbox{log} \, p_\theta(\x^u_t|\z_t)$, the training of sequential VAE generalizes from an unsupervised task to a supervised task that learns the model dynamics from past observed transitions and imputes the missing features. Given the pre-trained representation learning model, the policy is trained under a multi-stage reinforcement learning setting, where the representation provided by sequential VAE is taken as the input to the policy. The pseudocode for our proposed algorithm is presented in Appendix~\ref{app:algo}.

%% Experiments
%++++++++++++++++++++++++++++++++++++++++++++++++++
\section{Experiments}
\label{sec:exp}

% \begin{figure*}[b!]
% 	\centering
% 	\includegraphics[width=1.0\columnwidth]{latex/figures/recons_2.jpg}
% 	\caption{A sample trajectory for the \textit{bouncing ball}$^+$ domain generated from the initial policy, where both observation acquisition and control policies are randomly initialized.
% 	The three rows (top to bottom) correspond to: (1) the partially observable input selected by acquisition policy; (2) the ground-truth full observation; (3) reconstruction from \textit{Seq-PO-VAE}. %\sebastian{Is this true?}
% 	We notice that with the initial feature acquisition policy, the ball appears by chance. Quite a number of acquisitions are wasted without gaining useful information. }
% 	\label{fig:bouncing_ball_traj}
% \end{figure*}

We examine the characteristics of our proposed model in the following two experimental domains: a \textit{bouncing ball}$^+$ control task with high-dimensional image pixels as input, adapted from~\cite{fraccaro2017disentangled}; a \textit{sepsis} medical simulator fitted from real-world data~\cite{oberst2019counterfactual}.

%++++++++++++++++++++++++++++++++++++++++++++++++++
\paragraph{Baselines} 
For comparison, we mainly consider variants of the strong VAE baseline \textit{beta-VAE}~\cite{higgins2016beta}, which works on non-time-dependent data instances. For representing the missing features, we adopt the \textit{zero-imputing} method, proposed in~\cite{nazabal2018handling} over the unobserved features. Thus, we denote the VAE baseline as \textit{NonSeq-ZI}. We train the VAE with either the \textit{full} loss over the entire features, or the \textit{partial} loss which only applies to the observed features~\cite{MaTPHNZ19}. 
% Note that the sequential VAE trained on \textit{partial} loss reduces to a \textit{NonSeq-ZI} with the \textit{partial} loss, as the features to impute the observed part all stem from one single frame. % \sebastian{I don't understand the previous sentence.}
We also consider an \emph{end-to-end} baseline which does not employ pre-trained representation learning model. 
We denoted our proposed sequential VAE model for POMDPs as \textit{Seq-PO-VAE}. All the VAE-based approaches adopt an identical policy architecture. Detailed information on the model architecture is presented in appendix. We conduct all the experiments with 10 random seeds.

% \paragraph{Implementation}

% Specifically, for \textit{Bouncing Ball}$^+$ domain, the bottleneck size used for all VAEs are 32. The policy model is an LSTM-A3C, with the LSTM latent size to be 1024. For \textit{Sepsis} domain, we adopt a bottleneck size of 10 for all VAEs, and an LSTM latent size of 256. 
% For each task/method, the LSTM output is connected to a \textit{fc} layer with output size $|\mathcal{A}_f|$ for feature acquisition policy, another \textit{fc} layer with output size $|\mathcal{A}_c|$ for RL task policy, and another \textit{fc} with size $1$ to predict the state value. 
\vskip -2mm
\paragraph{Data Collection}
Pre-training the VAE models requires data generated by a non-random policy in order to incorporate abundant dynamics information. 
For both tasks, we collect a small scale dataset of 2000 trajectories, where half of the data is collected from a random policy and the the other half from a policy which better captures the state space that would be encountered by a learned model (e.g., by training a data collection policy end-to-end or using human generated trajectories). 
% Half of the data are derived from optimal trajectory\HY{do we need to explain and how shall we explain?} and the other half derived from random policy (for both feature acquisition and control).
The simple mixture of dataset works very well on both tasks without the need of further fine-tuning the VAEs.
% We demonstrate that such a simple mixture of dataset suffices to learn model dynamics and that our proposed sequential VAE model trained on the dataset performs very well on both tasks without further fine-tuning. 
We also create a testing set that consists of 2000 trajectories to evaluate the models. Additional details on data collection are available in Appendix~\ref{app:ball:data_collect} and \ref{app:sepsis:data_collect}.

%++++++++++++++++++++++++++++++++++++++++++++++++++
\subsection{Bouncing Ball$^+$}
\paragraph{Task Settings} We adapted the original \emph{bouncing ball} experiment presented in~\cite{fraccaro2017disentangled} by adding a navigation objective and control actions. Specifically, a ball moves in a 2D box and at each step, a binary image of size $32 \times 32$ showing the box and the ball is returned as the state. Initially, the ball appears at a random position in the upper left quadrant, and has a random velocity. The objective is to control the ball to reach a fixed target location set at $(5, 25)$. We incorporate five RL actions: a null action and four actions for changing the velocity of the ball in either the $x$ (horizontal) or $y$ (vertical) direction with a fixed scale: $\{\Delta V_x:\pm0.5, \, \Delta V_y:\pm0.5, \, null \}$.  The feature acquisition action is defined as selecting a subset from the four quadrants of image to observe. A reward of 1.0 is issued if the ball reaches its target location. Each episode runs up to 50 time steps.
% Figure~\ref{fig:bouncing_ball_traj} shows a sample trajectory.

\begin{wraptable}{r}{8cm}
\centering
\vspace{-15pt}
\caption{Missing feature imputing loss evaluated on \emph{Bouncing Ball}$^+$ and \emph{Sepsis}.  }
\label{table:missing_loss}
\vspace{2mm}
\begin{sc}
\scalebox{0.8}{
\begin{tabular}{lccr}
\toprule
\multirow{2}{*} {VAE model }    & Bouncing Ball$^+$ & Sepsis \\  & (NLL) & (MSE)  \\
\midrule
\multirow{2}{*} {NonSeq-ZI (partial)}   & 0.6504  & 0.8441   \\ & ($\pm$ 0.1391)  & ($\pm$0.0586) \\
\midrule
\multirow{2}{*} {NonSeq-ZI (full)}    & 0.0722 & 0.4839 \\  & ($\pm$ 0.0004) & ($\pm$ 0.0012)  \\
% NonSeq-PN (full)   &  & \\
% NonSeq-PN (partial)   &  & \\
\midrule
% \multirow{2}{*} {Seq-PO-VAE (w/o action)}   & - & - \\ & (-) & (-) \\
\multirow{2}{*} {Seq-PO-VAE (ours)}   & \textbf{0.0324} &  \textbf{0.1832} \\ & ($\pm$ 0.0082) & ($\pm$0.0158) \\
% \bottomrule
\end{tabular}
}
\end{sc}
\end{wraptable}

\begin{figure}[b!]
	\centering
	\includegraphics[width=.32\columnwidth]{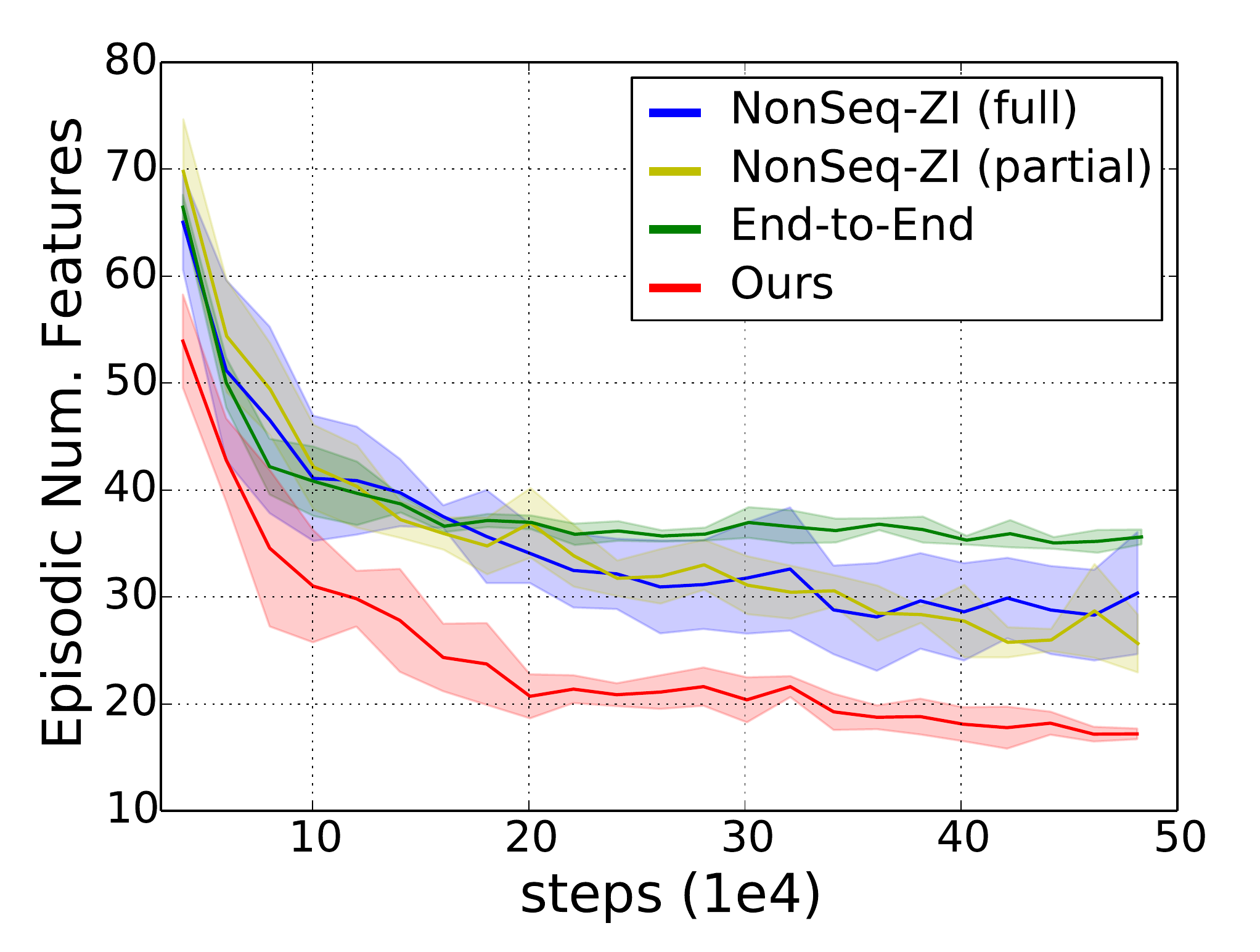}
	\includegraphics[width=.32\columnwidth]{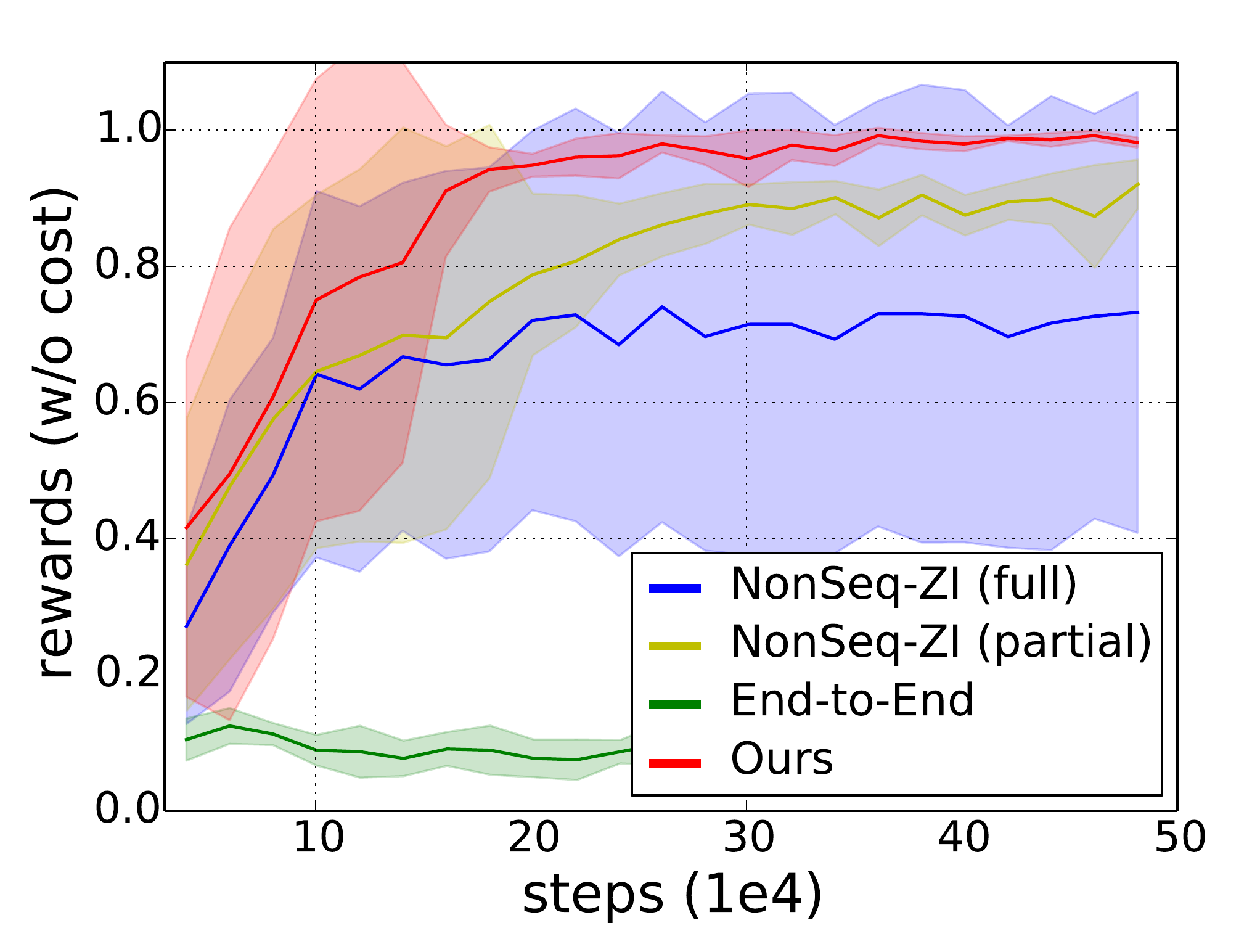}
	\includegraphics[width=.32\columnwidth]{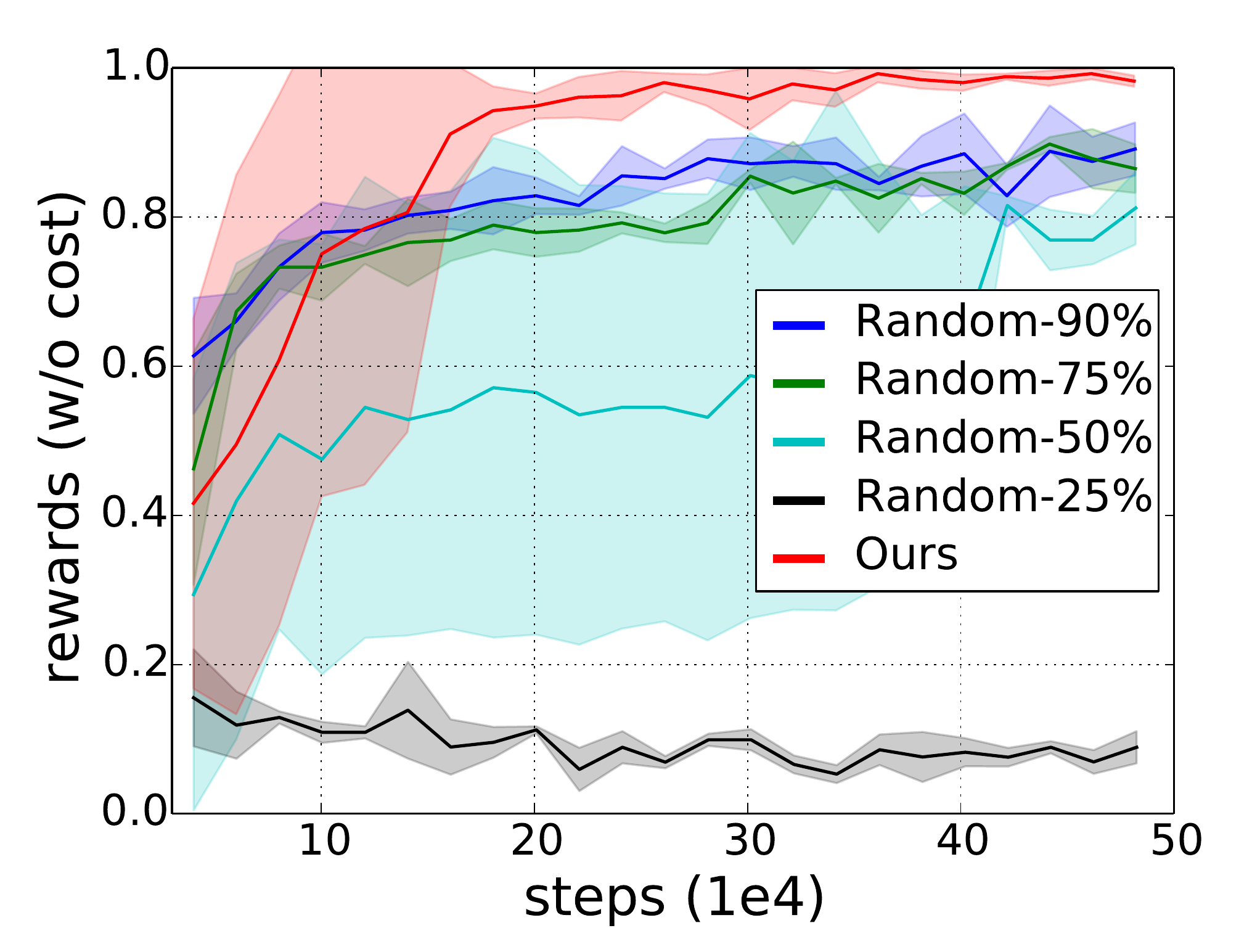}
	\vskip -0.1in
	\caption{Performance curves on the \emph{bouncing ball$^+$} domain: \textbf{a}: episodic number of observations acquired by the $\pi^f$; \textbf{b}: task rewards w/o cost.
	Our proposed method outperforms the non-sequential baselines in learning the task as well as acquiring less observations; \textbf{c}: Ablation study on \emph{bouncing ball$^+$} to illustrate the effect of learning the feature acquisition policy. 
% 	Each method is run with 3 random seeds. 
% 	Our proposed approach outperforms the random baseline significantly in terms of task performance.
	}
	\label{fig:bouncing_ball_perf}
\end{figure}

\begin{figure*}[t!]
	\centering
	\includegraphics[width=1.0\columnwidth]{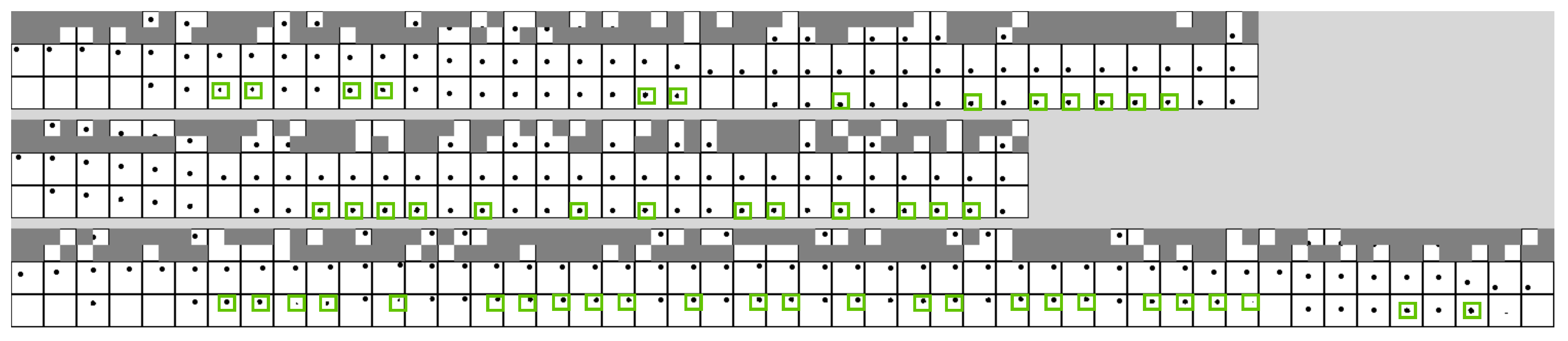}
	\vspace{-4mm}
	\caption{\textit{Seq-PO-VAE} reconstruction for the online trajectories upon convergence (better to view enlarged). Each block of three rows corresponds to the results for one trajectory. In each block, the three rows (top-down) correspond to: (1) the partially observable input selected by acquisition policy; (2) the ground-truth full observation; (3) reconstruction from \textit{Seq-PO-VAE}. The green boxes remark the frames where ball is not observed but our model could impute its location. Key takeaways: (1) our learned acquisition policy captures model dynamics ; (2) \textit{Seq-PO-VAE} effectively impute the missing features (i.e., ball can be reconstructed even when they are unobserved from consequent frames). 
% 	\CZ{Consider to use green boxes to mark the frames where the ball location is not observed in the training data but we can correctly infer the ball location.}
	}
% 	\vspace{-4mm}
	\label{fig:bouncing_ball_recons}
\end{figure*}
\vspace{-2mm}
\paragraph{Representation Learning Results} 
We evaluate the missing feature imputing performance of each VAE model in terms of negative log likelihood (NLL) and present results in Table~\ref{table:missing_loss}. We notice that our proposed model yields a significantly better imputing result than all the other baselines. This demonstrates that our proposed sequential VAE model can efficiently capture the environment dynamics and learn meaningful information over the missing features. Such efficiency is vital in determining both the acquisition and task policy training performance in AFA-POMDP, since both policies are conditioned on the VAE latent features. We also demonstrate sample trajectories reconstructed by different VAE models in the Appendix~\ref{sec:app:ball:impute}. The results show that our model learns to impute considerable amount of missing information given the partially observed sequence.

\vspace{-2mm}
\paragraph{Policy Training Results}
% We show that using our proposed \textit{Seq-PO-VAE} for policy training could lead to much better convergence compared to the baselines.
We evaluate the policy training performance in terms of episodic number of acquired observations and the task rewards (w/o cost). The results are presented in Figure~\ref{fig:bouncing_ball_perf} (a) and (b), respectively. First, we notice that the \emph{end-to-end} method fails to learn task skills under the given feature acquisition cost. However, the VAE-based representation learning methods manage to learn the navigation skill under the same cost setting. This verifies our assumption that representation learning plays a vital role in policy training under the AFA-POMDP scenario.
Furthermore, we also notice that the joint policies trained by \textit{Seq-PO-VAE} can develop the target navigation skill at a much faster pace than the non-sequential baselines. Our method also converges to a standard where much less feature acquisition is required to accomplish the task.

We also show that our proposed method can learn meaningful feature acquisition policies. To this end, we visualize three sampled trajectories upon convergence of training in Figure~\ref{fig:bouncing_ball_recons}. From the examples, we notice that our feature acquisition policy acquires meaningful features with a majority grasping the exact ball location. Thus, it demonstrates that the feature acquisition policy adapts to the dynamics of the problem and learns to acquire meaningful features. 
% We also show a case study that our method can learn a near optimal acquisition policy for a reduced task, in Appendix.
We also show the actively learned feature acquisition policy works better than random acquisition. From the results in Figure~\ref{fig:bouncing_ball_perf} (c), our method converges to substantially better standard than random policies with considerably high selection probabilities.

\begin{figure*}[b!]
	\centering
	\vspace{-4mm}
	\includegraphics[width=.32\columnwidth]{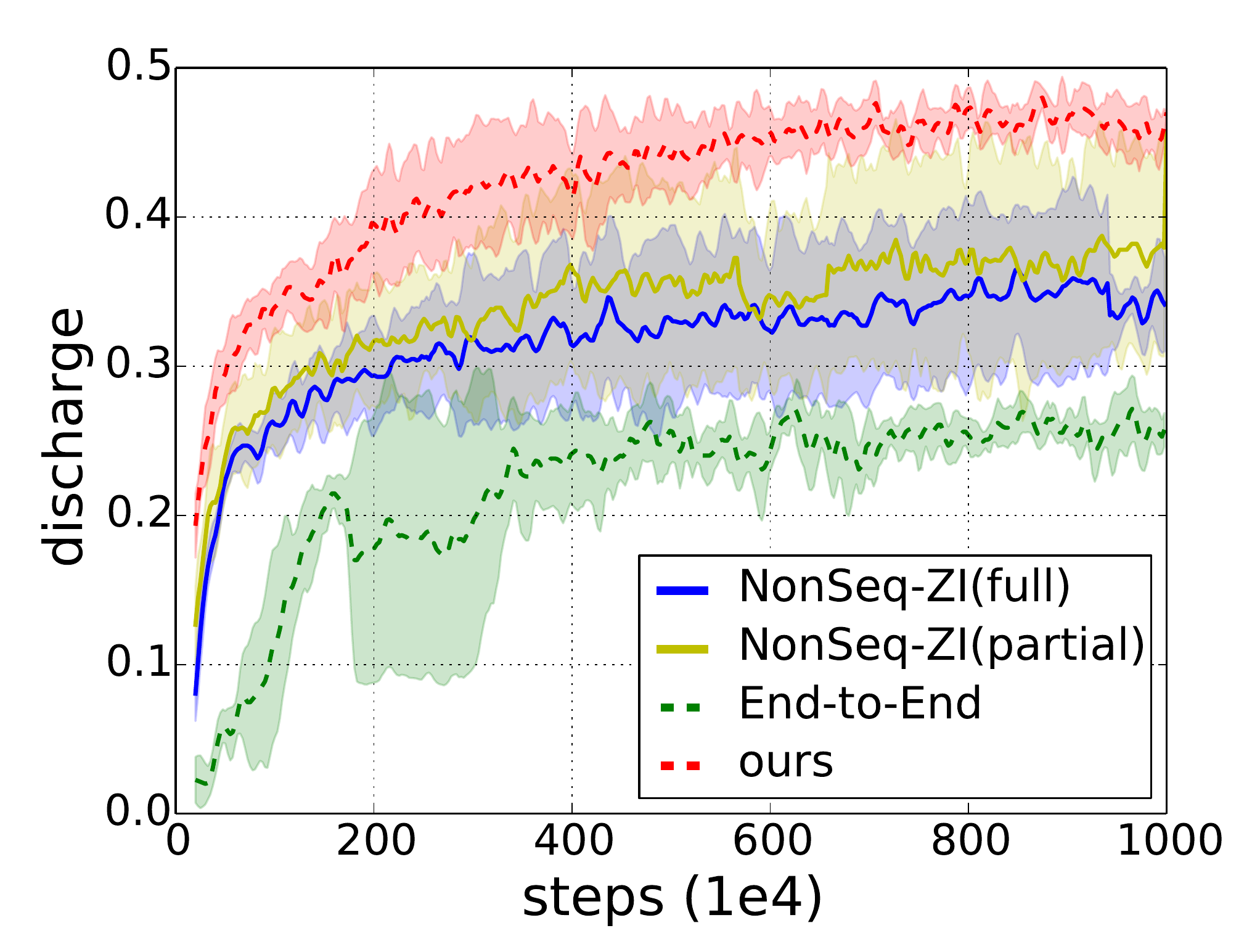}\hspace{.012\columnwidth}
	\includegraphics[width=.32\columnwidth]{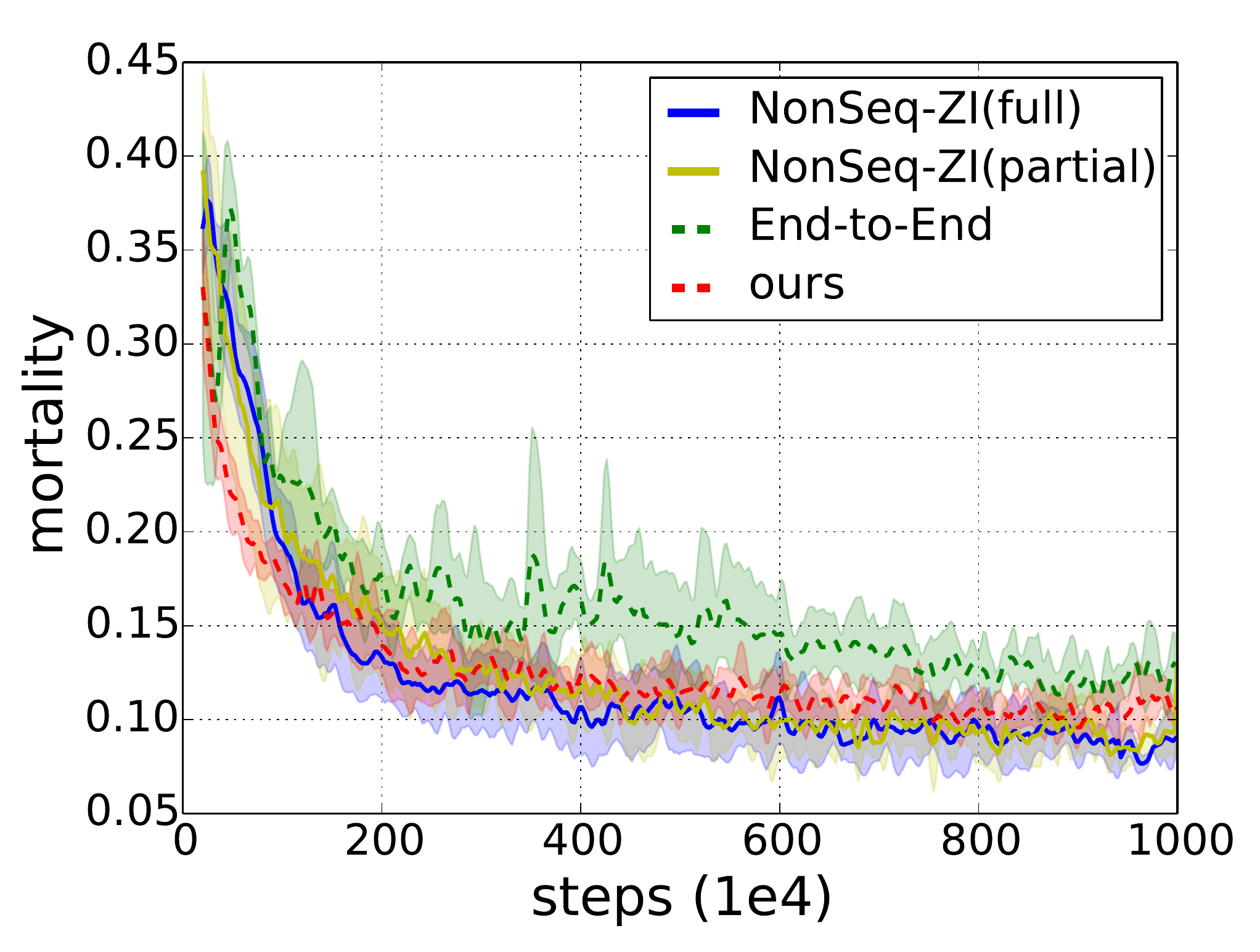}\hspace{.012\columnwidth}
	\includegraphics[width=.32\columnwidth]{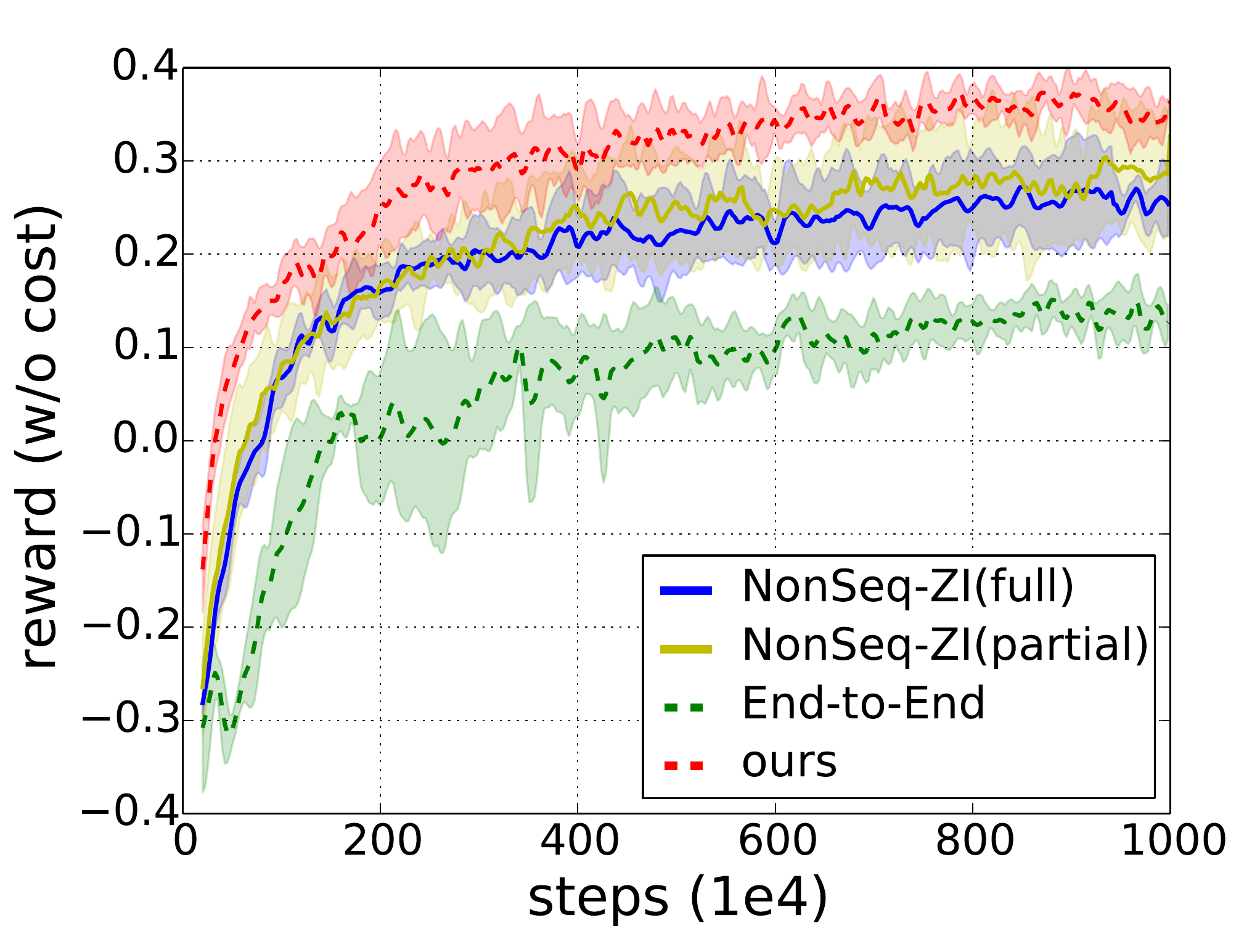}
	\vskip -0.05in
	\caption{Performance curves in terms of discharge rate, mortality rate and reward (w/o cost) for the compared approaches on \emph{Sepsis}. The curves are derived under cost value of 0.01. Overall, our method converges to treatment policy with substantially better reward compared to the baselines.
	}
	\label{fig:sepsis_performance}
\end{figure*}
\subsection{Sepsis Medical Simulator}
\paragraph{Task Settings} Our second evaluation domain adopts a medical simulator for treating sepsis among ICU patients, proposed in~\cite{oberst2019counterfactual}. Overall, the task is to learn to apply three \emph{treatment} actions to the patient, i.e, $\{$\emph{antibiotic}, \emph{ventilation}, \emph{vasopressors}$\}$.  The state space consists of 8 features: 3 of them indicate the current \emph{treatment} state for the patient; 4 of them are the \emph{measurement} states over \emph{heart rate}, \emph{sysBP rate}, \emph{percoxyg state} and \emph{glucose level}; the rest is an index specifying the patent's \emph{diabetes} condition. The feature acquisition policy learns to actively select the \emph{measurement} features. Each episode runs for up to 30 steps. The patient will be discharged if his/her \emph{measurement} states all return to normal values. An episode terminates upon mortality or discharge, with a reward $-1.0$ or $1.0$.

\vspace{-2mm}
\paragraph{Representation Learning Results}
We evaluate the missing feature imputing performance for each VAE model on the testing dataset. The loss is evaluated in terms of MSE and we present the results in Table~\ref{table:missing_loss}. Overall, our model results in the lowest MSE loss. Again this result shows that our proposed sequential VAE model could learn reasonable imputation over missing features with the learned model dynamics on tasks with stochastic transitions.

\vspace{-2mm}
\paragraph{Policy Training Results} 
We show the policy training results for \emph{Sepsis} in Figure~\ref{fig:sepsis_performance}. Overall, our proposed method results in substantially better task reward compared to all baselines. Note that the performance of discharge rate for our method increases significantly faster than baseline approaches, which shows that the model can quickly learn to apply appropriate treatment actions and thus be trained in a much more sample efficient way. Moreover, our method also converges to substantially better values than the baselines. Upon convergence, it outperforms the best non-sequential VAE baseline with a gap of $> 5\%$ for discharge rate. 
For all the evaluation metrics, we notice that VAE-based representation learning models outperform the end-to-end baseline by significant margins. This indicates that efficient representation learning is crucial to determine the effect of agent's policy training practice. The result also reveals that learning to impute missing features has the potential to contribute greatly to improve the policy training performance for AFA-POMDP.

%+++++++++++++++++++++++++++++++++++++++++++++
\subsection{Ablation Study}
In this section, we present an ablation study on the \emph{Sepsis} medical domain. 

\begin{wrapfigure}{r}{0.4\columnwidth}
\centering
\vspace{-15pt}
\includegraphics[width=0.4\columnwidth]{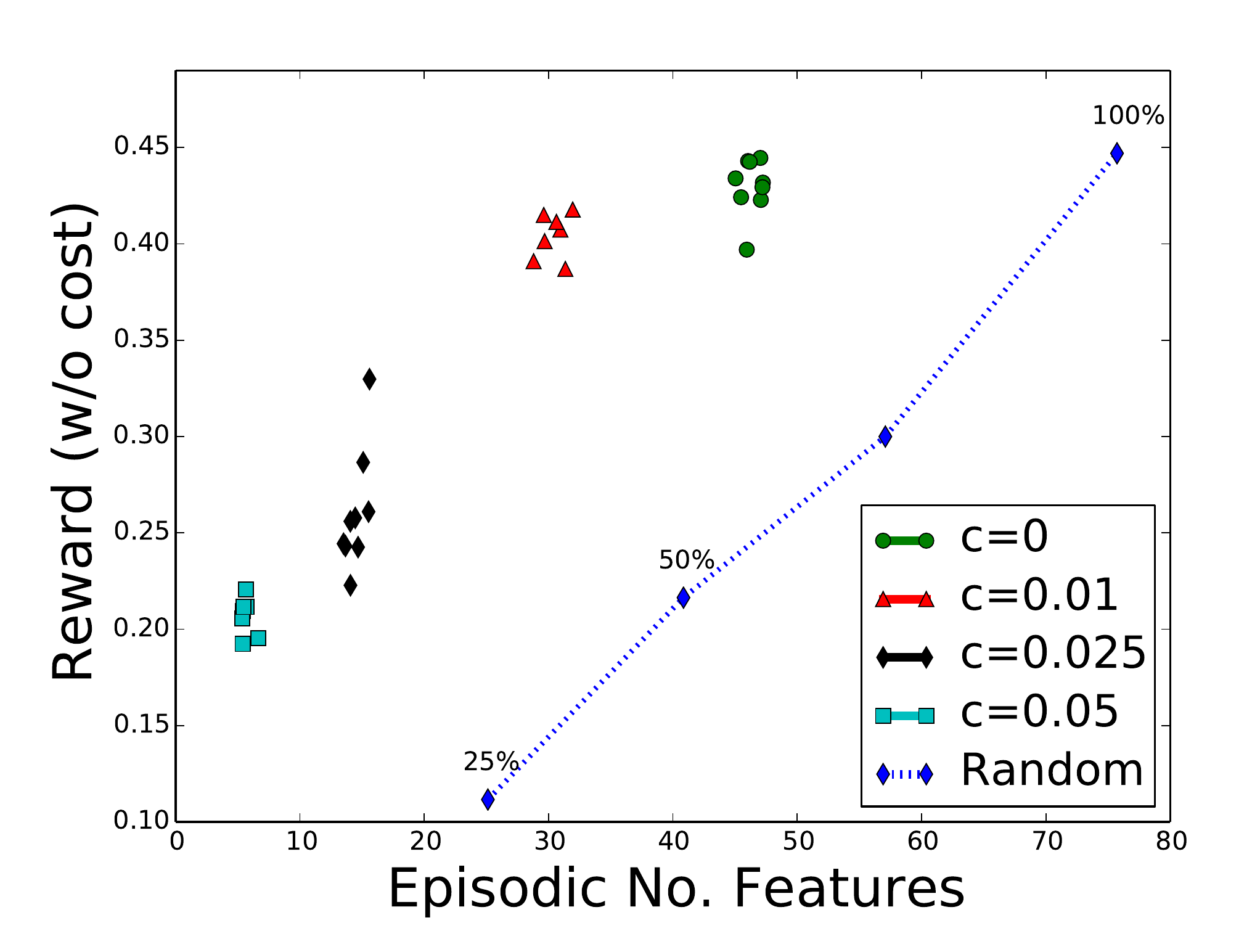}
\vspace{-4mm}
\caption{
Active feature acquisition (under different cost values) vs. random feature acquisition.
}\label{fig:sepsis_cost}
\vspace{-15pt}

\end{wrapfigure}

\vspace{-2mm}
\paragraph{Efficacy of Active Feature Acquisition}
We study the effect of actively learning sequential feature acquisition strategy with RL. To this end, we compare our method with a baseline that randomly acquires features. We evaluate our method under different cost values,
% We show the results for  \textit{Bouncing Ball}$^+$ in Figure~\ref{fig:bouncing_ball_ablation}. We notice that without an actively learned feature acquisition policy, the task performance for the random baseline converges to an inferior standard. Our actively learned feature acquisition policy  outperforms random baselines by a great performance margin.  
% Moreover, we notice that learning the feature acquisition policy with RL can significantly benefit the reduction of feature acquisition cost. While the random acquisition policy always converge to a fixed standard in terms number of acquired observations per episode regardless of the cost for acquisition, our proposed method can result in diverse acquisition strategies by setting the acquisition cost. Overall, our learned acquisition policy ensures desirable task performance by acquiring $< 20$ features. Such value is much lower than that for the \textit{Random}-$25\%$ baseline, which is $> 30$.  
% Therefore, we could conclude that performing active feature acquisition on RL tasks could help to reduce the feature acquisition cost with great efficacy.
and the results are shown in Figure~\ref{fig:sepsis_cost}. From the results, we notice that there is a clear cost-performance trade-off, i.e., a higher feature acquisition cost results in feature acquisition policies that obtain fewer observations, with a sacrifice of task performance. Overall, our acquisition method results in significantly better task performance than the random acquisition baselines. Noticeably, with the learned active feature acquisition strategy, we acquire only about half of the total number of features (refer to the x-value derived by \textit{Random}-$100\%$) to obtain comparable task performance. 
% Also, we notice that the specified cost has a very clear impact on the final task performance, i.e., the number of acquired features per episode decreases significantly as the cost increases. Thereby, our proposed solution can promisingly compute feature acquisition policies that meet different budgets. 

\vspace{-2mm}
\paragraph{Impact on Total Acquisition Cost}
\begin{wrapfigure}{r}{0.4\columnwidth}
\centering
\vspace{-15pt}
\includegraphics[width=.4\columnwidth]{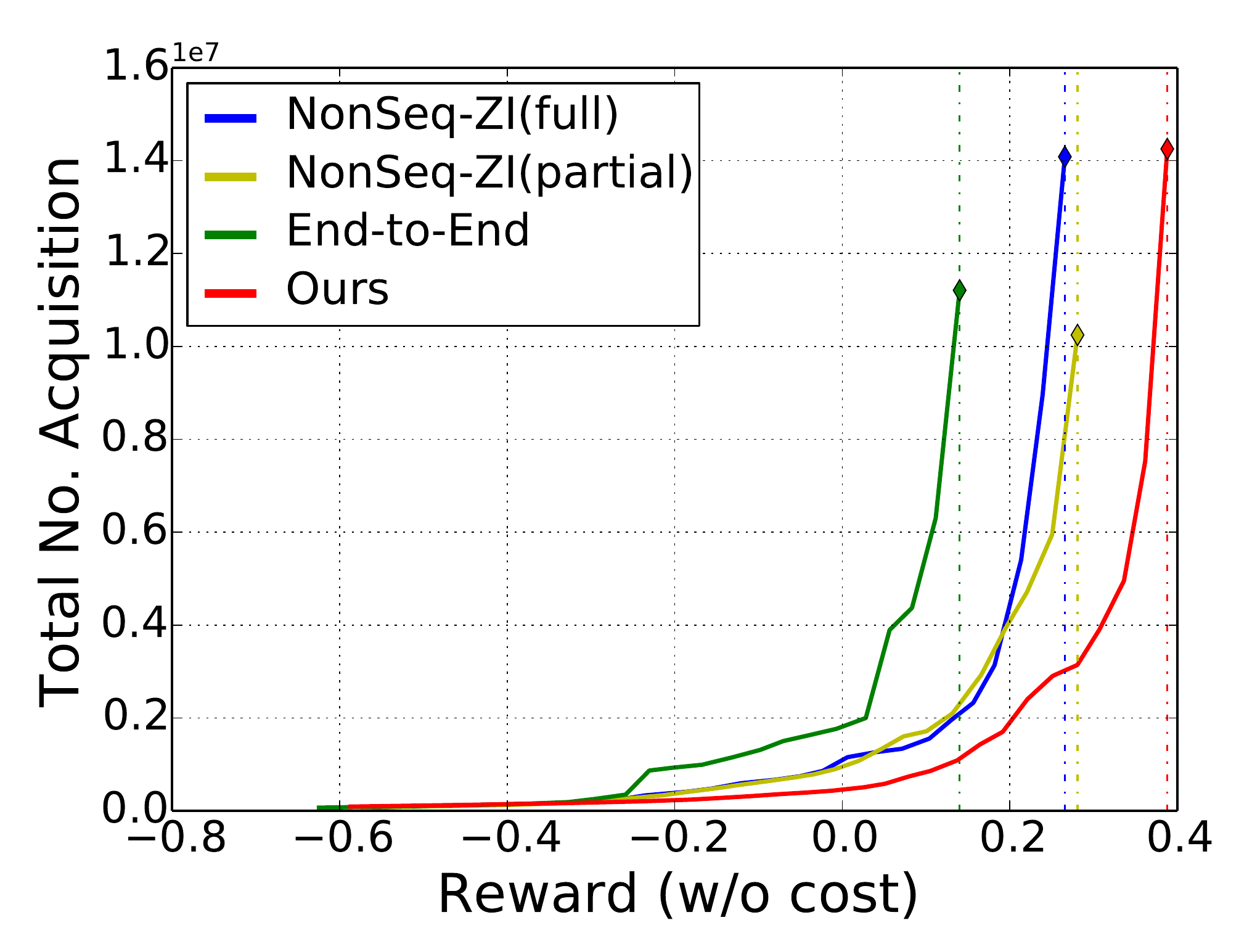}
\vspace{-4mm}
\caption{
Total feature acquisition cost consumed by different approaches.
}\label{fig:sepsis_rew_cost}
\vspace{-15pt}
\end{wrapfigure} 
For different representation learning methods, we also investigate the total number of features acquired at different stage of training. The results are shown in Figure~\ref{fig:sepsis_rew_cost}. As expected, to obtain better task policies, the models need to take longer training steps and thus the total feature acquisition cost would increases accordingly. We notice that policies trained by our method result in the highest convergent task performance (max x-value). Given a certain performance level (same x-value), our method consumes substantially less total feature acquisition cost (y-value) than the others. We also notice that the overall feature acquisition cost increases with a near exponential trend. Overall, conducting policy training for AFA-POMDP with our proposed representation learning method could lead to subsequent reduce in total feature acquisition cost compared to the baseline methods.
% Such findings also pinpoint an important future research direction, which is to employ advanced exploration approaches to further reduce the total feature acquisition cost.

% \subsection{Pre-training versus End-to-end}

\vspace{-2mm}

%% Conclusion
\section{Conclusion}
In this paper, we present a novel AFA-POMDP framework where the task policy and the active feature acquisition policy are learned under a unified formalism. Our method incorporates a model-based representation learning attempt, where a sequential VAE model is trained to impute missing features via learning model dynamics and thus offer high quality representations to facilitate the joint policy training under partial observability. Our proposed model, by efficiently synthesizing the sequential information and imputing missing features, can significantly outperform conventional representation learning baselines and leads to policy training with significantly better sample efficiency and obtained solutions. Future work may investigate more cost-sensitive application domains 
to apply our proposed method. Another promising direction is to study hierarchical representations for the feature acquisition actions in AFA-POMDPs, which could likely alleviate issues caused by the large action space.

%% Broader impact
\section*{Broader Impact} %this can be on page 9

When deploying machine learning models in real-world applications, the fundamental assumption that the features used during training are always readily available during the deployment phase does not necessarily hold. Our proposed approach could relax such assumptions and enable machine learning models to be used in a broader range of application domains. 

Our paper also opens an interesting new research direction for active learning, which extends the conventional instance-wise non-time-dependent active feature acquisition task to a more challenging time-dependent sequential decision making task. This task has important implications for real-life applications, such as healthcare and education. We demonstrate the great potential and practicality of deriving cost-sensitive decision making strategies with active learning. 
% It has great implication in deriving cost-efficient task solutions.

Considering that learning and applying the models is problem specific, it is unlikely that our method can equally benefit all possible application scenarios.
We also fully acknowledge the existence of risk in applying our model in sensitive and high risk domains, e.g., healthcare, and bias if the model itself or the used representations are trained on biased data. 
In high risk settings, human supervision of the proposed model might be desired and the model could mainly be used for decision support.
However, there are still many practical scenarios that could satisfy our model assumption and are less sensitive.
We believe our paper has the potential to trigger many follow-up works studying novel models, theoretical aspects, and the application in different domains. It also provides opportunities for future works to align the data acquisition and task policy training process with risk, fairness and privacy concerns.

\bibliographystyle{abbrv}
\bibliography{refs}

\appendix
\clearpage

{\huge \bfseries Appendix}\\[4mm]
\noindent  This supplementary material is organized as follows. First, we present the detailed algorithm for our proposed \emph{active feature acquisition POMDP} (AFA-POMDP). Then we present additional experiment details on the \emph{BouncingBall+} task and the \emph{Sepsis} task. For each task, we present the task specifications, implementation details and additional evaluation results. Lastly, we present a case study that investigates the efficiency of our proposed sequential representation model when trained with data under different levels of observability.

\begin{appendices}

%% Algorithm
\section{AFA-POMDP Algorithm}
\label{app:algo}
% Our proposed algorithm \emph{active feature acquisition POMDP} (AFA-POMDP) is presented in Algorithm~\ref{alg:SAFA-RL}.

\begin{algorithm}[H]
   \caption{AFA-POMDP Algorithm}
   \label{alg:SAFA-RL}
\begin{algorithmic}[1]
   \STATE {\bfseries Input:} learning rate $\alpha > 0$, dataset $\mathcal{D}$
   \STATE {\bfseries Initialize} RL policy $\pi_f, \pi_c$, VAE parameters $\theta$, $\phi$. % \sebastian{According to above, aren't the VAE parameters $\theta$?}
   \STATE {\bfseries Train} VAE on dataset $\mathcal{D}$ using Eq (5).
   \WHILE{not converged}
   \STATE Reset the environment.
   \STATE Initialize null observation $\x_1^p = \O$, feature acquisition action $\veca^f_0$ and control action $\veca^c_0$. 
%   \STATE Sample an initial feat.\ acqu.\ action $\veca_0^f \sim \zeta (\O)$. \sebastian{Unclear how $\theta^o$ relates to the policies introduced in the text.}
   \FOR{$i=1$ {\bfseries to} $T$}
   \STATE Compute representation with VAE: $\vecb_t=q_\phi(\x^p_{\leq t}, \veca_{< t})$.
   \STATE Sample a feature acquisition action $\veca_t^f \sim \pi_f(\vecb_t)$ and a control action $ \veca_t^c \sim \pi_c(\vecb_t)$.
   \STATE Step the environment and receive partial features, reward and terminal signal: $\x^p_{t+1}, r_t, \textnormal{term} \sim \mbox{env}(\veca_{t}^f, \veca_{t}^c)$
   \STATE Compute cost $c_t=\sum_i c\cdot\mathbb{I}(\veca_{t}^{f(i)})$.
   %\sebastian{Should we interact with the environment through $\veca^f$ and $\veca^c$ jointly?}
   \STATE Save the transitions $\{\vecb_t, \veca_t^f, \veca_t^c, r_t, c_t, \textnormal{term}\}$.
%   \STATE{}
  \IF{$\textnormal{term}$}
  \STATE break
  \ENDIF
   \ENDFOR
   \STATE Update $\pi_f$, $\pi_c$ using the saved transitions with an RL algorithm using learning rate $\alpha$.
   \ENDWHILE
\end{algorithmic}
\end{algorithm}

%% Bouncing Ball
% \input{appendix_bouncing_ball.tex}
%+++++++++++++++++++++++++++++++++++
\section{Bouncing Ball$^+$}
\label{appendix:ball}

\subsection{Task Specifications}
The task consists of a ball moving in a 2D box of size $32 \times 32$ pixels.
The radius of the ball equals to $2$ pixels.
At each step, a binary image is returned as an observation of the MDP state.
At the beginning of every episode, the ball starts at a random position in the \emph{upper left} quadrant (sampled uniformly).
The initial velocity of the ball is randomly defined as follows: $\vec{v}=[V_x,V_y]=4 \cdot \tilde{\vec{v}} / \| \tilde{\vec{v}} \|$, where the x- and y-component of $\tilde{\vec{v}}$ are sampled uniformly from the interval $[-0.5,0.5]$.
% \sebastian{I am not exactly sure about the velocity. Is it correct that $\beta=4$ and that there is another $4$ in front of the uniform distribution? Also this is inconsistent with the main text in terms of notation? Should we just say the following: $\vec{v}=[V_x,V_y]=4 \cdot \tilde{\vec{v}} / \| \tilde{\vec{v}} \|$, where the x- and y-component of $\tilde{\vec{v}}$ are sampled uniformly from the interval $[-0.5,0.5]$.}
There is a navigation target set at $(5,25)$ pixels, which is in the \emph{lower left} quadrant. The navigation is considered to be successful if the ball reaches the specified target location within a threshold of 1 pixel along both x/y-axis.

The action spaces is defined as follows.
There are five task actions $\mathcal{A}^c$:
\begin{itemize}
    \item Increase velocity leftwards, i.e., change $V_x$ by $-0.5$
    \item Increase velocity rightwards, i.e., change $V_x$ by $+0.5$
    \item Increase velocity downwards, i.e., change $V_y$ by $+0.5$
    \item Increase velocity upwards, i.e., change $V_y$ by $-0.5$
    \item Keep velocities unchanged
\end{itemize}
%$=\{\Delta V_x:\pm0.5, \, \Delta V_y:\pm0.5, \, \textnormal{null} \}$, where the effective action changes the velocity along either x/y-axis at a fixed scale.
The maximum velocity along the x/y-axis is 5.0. The velocity will stay unchanged if it exceeds this threshold. The feature acquisition action $\veca^f \in \mathcal{A}^f$ is specified as acquiring the observation of a subset of the quadrants (this also includes acquiring the observation of all 4 quadrants).
Thus, the agent can acquire $0-4$ quadrants to observe.
Each episode runs up to 50 steps.
The episode terminates if the agent reaches the target location.

\subsection{Implementation Details}
For all the baseline methods, \emph{Zero-Imputing}~\cite{nazabal2018handling} is adopted to fill in missing features with a fixed value of 0.5. 

\paragraph{End-to-End} The end-to-end model first processes the imputed image by 2 \emph{convolutional} layers with filter sizes of 16 and 32, respectively. Each \emph{convolutional} layer is followed by a \emph{ReLU} activation function. Then the output is passed to a \emph{fully connected} layer of size 1024. The weights for the \emph{fully connected} layer are initialized by \emph{orthogonal weights initialization} and the biases are initialized as zeros.

\paragraph{NonSeq-ZI} The non-sequential VAE models first process the imputed image by 2 \emph{convolutional} layers with filter sizes of 32 and 64, respectively. Each \emph{convolutional} layer is followed by a \emph{ReLU} activation function. Then the output passes through a \emph{fully connected} layer of size 256, followed by two additional \emph{fully connected} layers of size $32$ to generate the mean and variance of a Gaussian distribution.
To decode an image, the sampled code first passes through a \emph{fully connected} layer with size $256$, followed by 3 \emph{deconvolutional} layers with filters of 32, 32, and $nc$ and strides of 2, 2 and 1, respectively, where $nc$ is the \emph{channel} size that equals to 2 for the binary image.
% \sebastian{Unclear, how the code comes into the picture.}
There are two variants for \emph{NonSeq-ZI}: one employs the \emph{partial} loss that is only computed for the observed features; the other employs the \emph{full} loss that is computed for all the features, i.e., the ground-truth image with full observation is employed as the target to train the model to impute the missing features. The hyperparameters for training \emph{NonSeq-ZI} are summarized in Table~\ref{table:ball_vae_params}. 

\begin{table}[b!]
  \caption{Hyperparameter settings for training VAE models on the \emph{Bouncing Ball}$^+$ dataset. }
\label{table:ball_vae_params}
  \centering
  \begin{tabular}{rcccc}
    \toprule
    & \multicolumn{4}{c}{Hyperparameters}\\\cmidrule{2-5}
     & $\beta$ (KL weight) & KL reduction & Loss reduction & learning rate \\
    \midrule
    NonSeq-ZI (partial) & 1.0 & sum & sum & 1e-4 \\
    NonSeq-ZI (full) & 1.0 & sum & sum & 1e-4  \\
    Seq-PO-VAE (ours) & 1.0 & sum & sum & 5e-4  \\
    \bottomrule
  \end{tabular}
\end{table}

\paragraph{Seq-PO-VAE (ours)}
At each step, the \emph{Seq-PO-VAE} takes an imputed image and an action vector of size 9 as input. The imputed image is processed by 3 \emph{convolutional} layers with filter size 32 and stride 2. Each \emph{convolutional} layer employs \emph{ReLU} as its activation function. Then the output passes through a \emph{fully connected} layer of size 32 to generate a latent representation for the image $\f_x$. The action vector passes through a \emph{fully connected} layer of size 32 to generate a latent representation for the action $\f_a$. Then the image and action features are concatenated and augmented to form a  feature vector $\f_c = [\f_x,\, \f_a,\, \f_x*\f_a]$, where $[\cdot]$ denotes \emph{concatenation} of features.
Then $\f_c$ is fed to \emph{fully connected} projection layers of size 64 and 32, respectively. The output is then fed to an \emph{LSTM} module, with latent size of 32. The output $\h_t$ of \emph{LSTM} is passed to two independent \emph{fully connected} layers of size 32 for each to generate the mean and variance for the Gaussian distribution filtered from the sequential inputs. To decode an image, the model adopts \emph{deconvolutional} layers that are identical to those for \emph{NonSeq-ZI}. The hyperparameters for training \emph{Seq-PO-VAE} are shown in Table~\ref{table:ball_vae_params}.

\paragraph{LSTM-A3C} We adopt LSTM-A3C~\cite{mnih2016asynchronous} to train the RL policy. The policy takes the features derived from the representation learning module as input. For the VAE-based methods, the input features are passed through a \emph{fully connected} layer of size 1024. Then the features are fed to an \emph{LSTM} with 1024 units. The output of the \emph{LSTM} is fed to three independent \emph{fully connected} layers to generate the estimations for value, task policy and feature acquisition policy. We adopt \emph{normalized column} initialization for all the \emph{fully connected} layers and the biases for the \emph{LSTM} module are set to zero.

\subsection{Data Collection}
\label{app:ball:data_collect}
To train the VAEs, we prepare a training set that consists of 2000 trajectories. Half of the trajectories are derived from a random policy and the other half is derived from a policy learned from an end-to-end method. To train the end-to-end method, we employ a cost of 0.01 over the first 2m steps and then increase it to 0.02 for the following 0.5m steps. All the VAE models are evaluated on a test dataset that has identical size and data distribution as the training dataset.
We present the best achieved task performance of the data collection policy (\emph{End-to-End}) and our representation learning approach in Table~\ref{table:bouncingball_task_obs}. We notice that our proposed method, by employing an advanced representation model, leads to a significantly better feature acquisition policy than \emph{End-to-End} (smaller number of observations while achieving similar or better reward).
% \sebastian{Is the remark I added in brackets correct? Can we add reward information?}  \HY{Yes correct. The task reward for both cases is 1.0}

\begin{table}[h!]
  \centering
  \caption{ Task performance for the data collection policy and our proposed method on \emph{Bouncing Ball}$^+$.}
  \label{table:bouncingball_task_obs}
  \begin{tabular}{rcc}
    \toprule
    & \multicolumn{2}{c}{Model} \\\cmidrule{2-3}
     & End-to-End & Ours \\
    \midrule
    Average \# of observations per episode & 17.94 & \textbf{8.24} \\
    Task reward & 1.0 & 1.0 \\
    \bottomrule
  \end{tabular}
\end{table}

\clearpage
\subsection{Imputing Missing Features via Learning Model Dynamics}
\label{sec:app:ball:impute}
We present an illustrative example to demonstrate the process of imputing missing features and the role of learning model dynamics.
To this end, we collect trajectories under an \emph{End-to-End} policy (the choice of the underlying RL policy is not that important since we just want to derive some trajectory samples for the VAE models to reconstruct) and use different VAE models to impute the observations. 
% \sebastian{What does identical mean here? Do you mean that they are from the same distribution as the one the VAE was trained on? Please clarify in the text.}

\begin{figure}[h!]
	\centering
% 	\vskip 0.1in
	\includegraphics[width=.67\columnwidth]{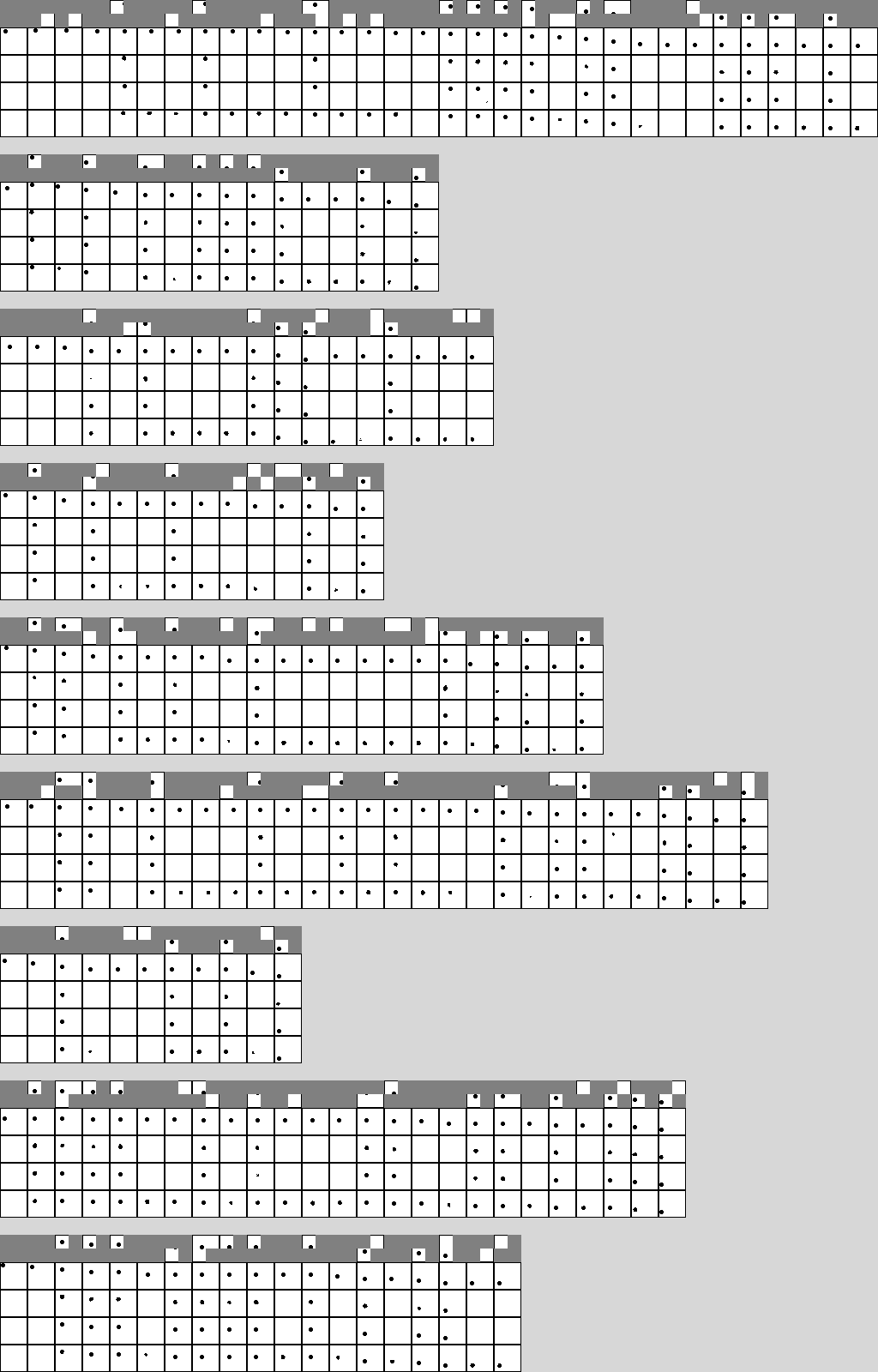}
	\caption{Imputation results for different VAE models. We select 9 trajectories obtained from the trained \emph{End-to-End} policy. Each block corresponds to the results for one trajectory (better to view enlarged). The five rows in one block are (top-down): (1) partial observations acquired by the agent; (2) ground-truth image with full observation; (3) Imputation by \emph{NonSeq-ZI (partial)}; (4) Imputation by \emph{NonSeq-ZI (full)}; (5) Imputation by \emph{Seq-PO-VAE (ours)}. 
	Our model can often successfully predict the balls location even if it is not present in the acquired observation. Hence it successfully employs its learned knowledge of the dynamics.
	In contrast, the non-sequential model (obviously) fails to predict the balls location when the ball is not present in the observation.}
	\label{fig:ball_impute_traj}
\end{figure}

From the results presented in Figure~\ref{fig:ball_impute_traj}, we observe that under the partially observable setting with missing features, the latent representation derived from our proposed method provides abundant information as compared to only using information from a single time step and thereby offers significant benefit for the policy model to learn to acquire meaningful features/gain task reward.

\subsection{Investigation on Cost-Performance Trade-off}
\begin{figure}[h!]
\centering
\vskip -0.1in
\subfigure[End-to-End]{\label{fig:ball_cnn}
	\includegraphics[width=.3\columnwidth]{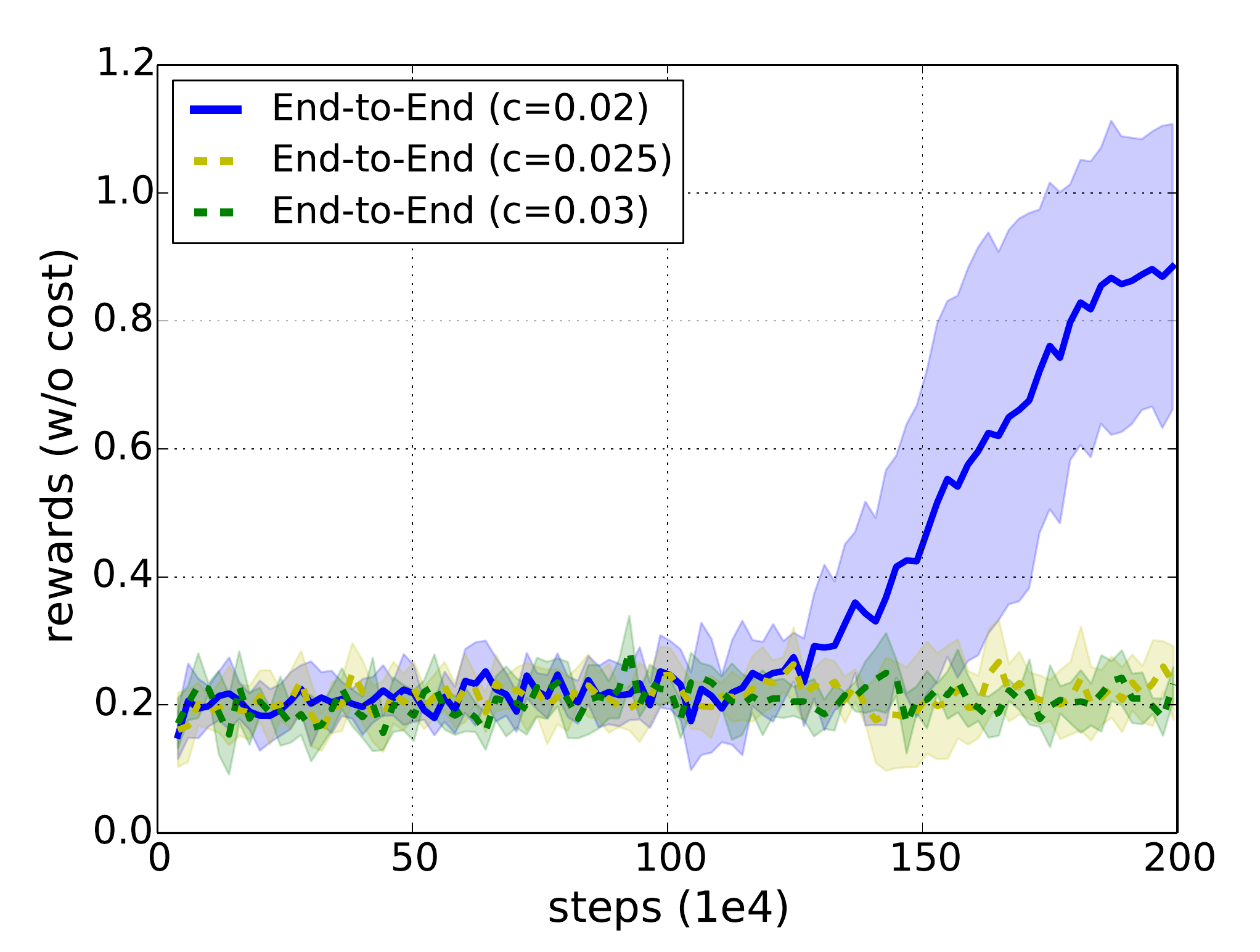} \hspace{4mm}
	\includegraphics[width=.3\columnwidth]{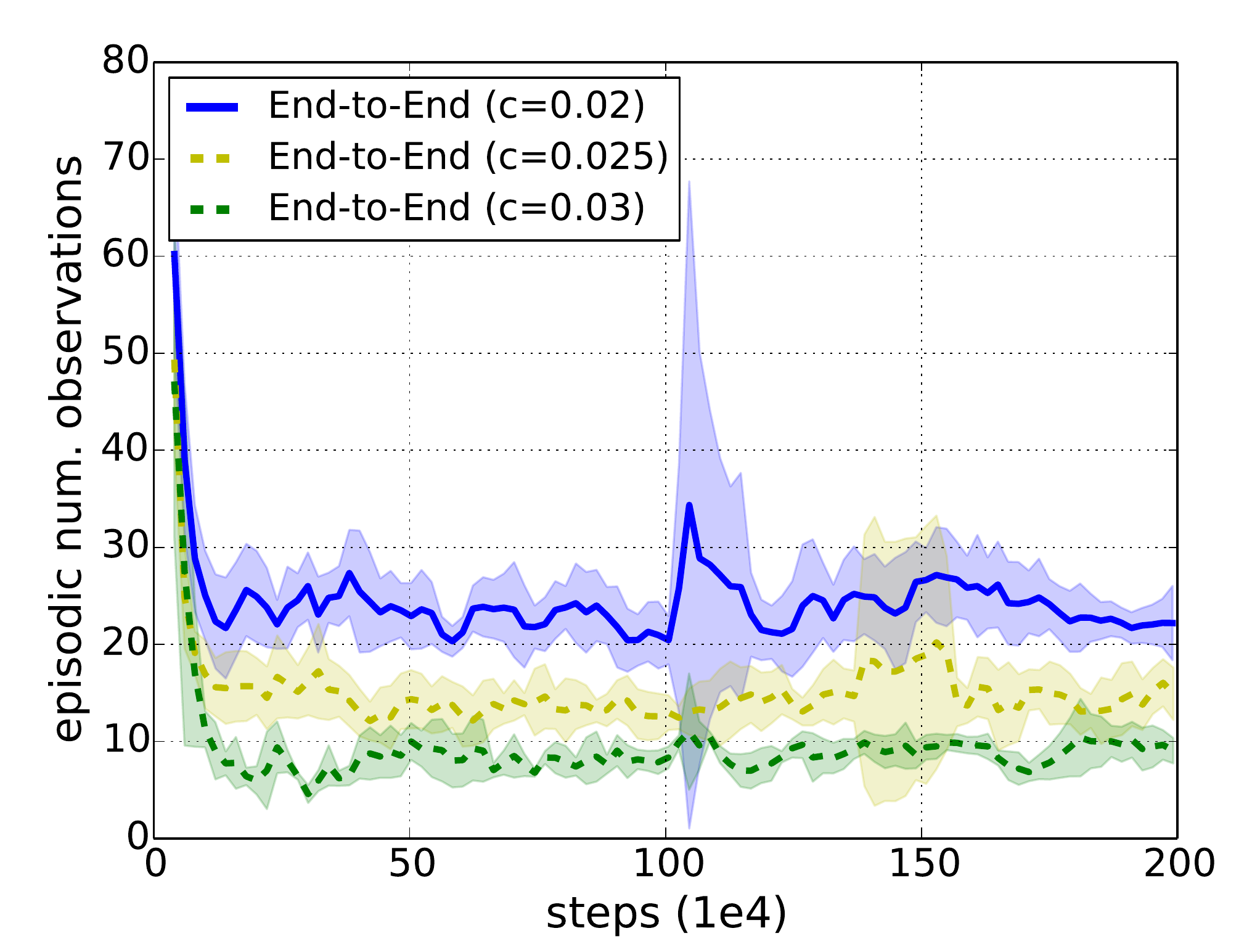}
}\\
\vskip -0.08in
\subfigure[NonSeq-ZI (full)]{\label{fig:ball_nonseq_full}
\includegraphics[width=0.3\textwidth]{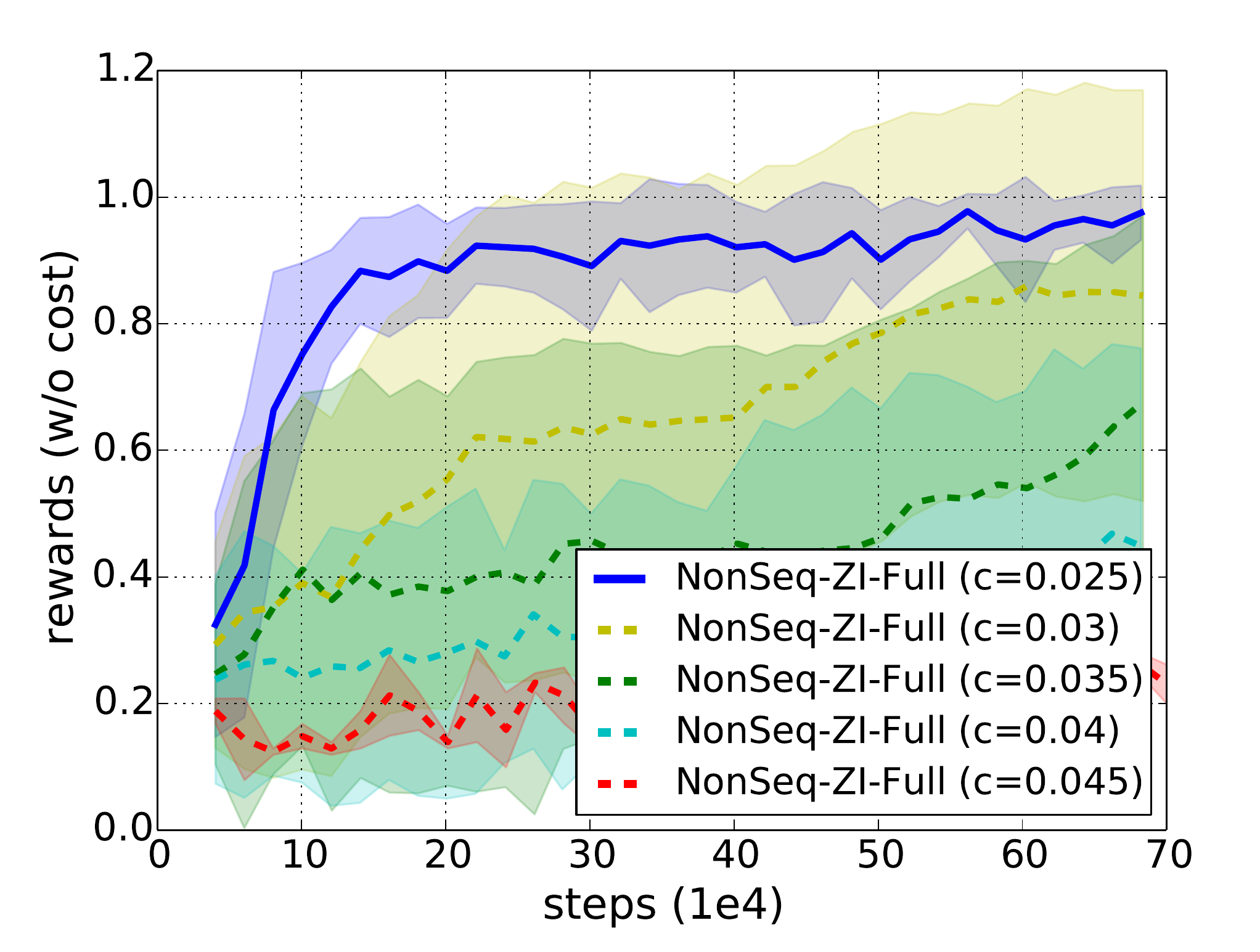} \hspace{4mm}
\includegraphics[width=0.3\textwidth]{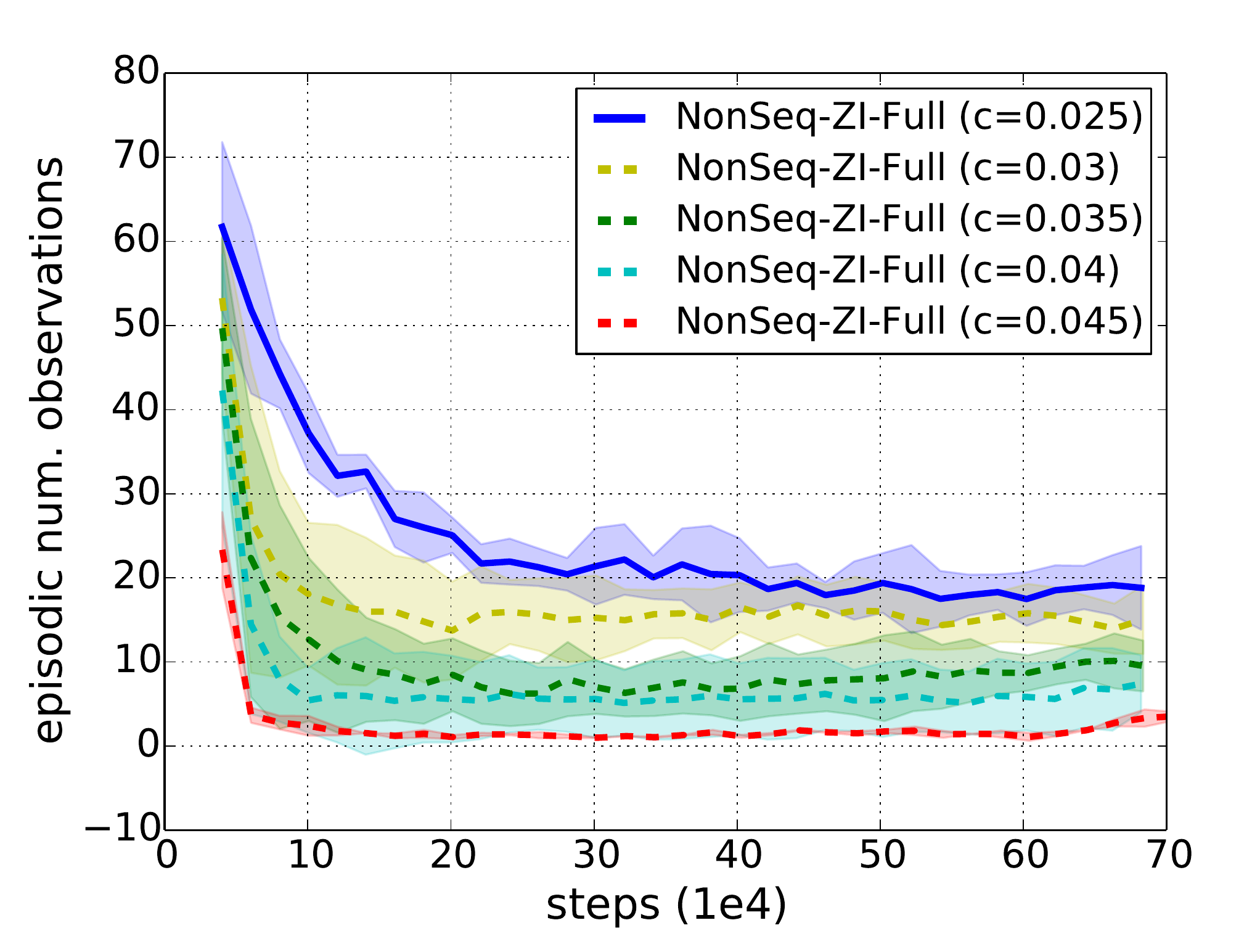}
}\\
\vskip -0.08in
\subfigure[NonSeq-ZI (partial)]{\label{fig:ball_nonseq_partial}
\includegraphics[width=0.3\textwidth]{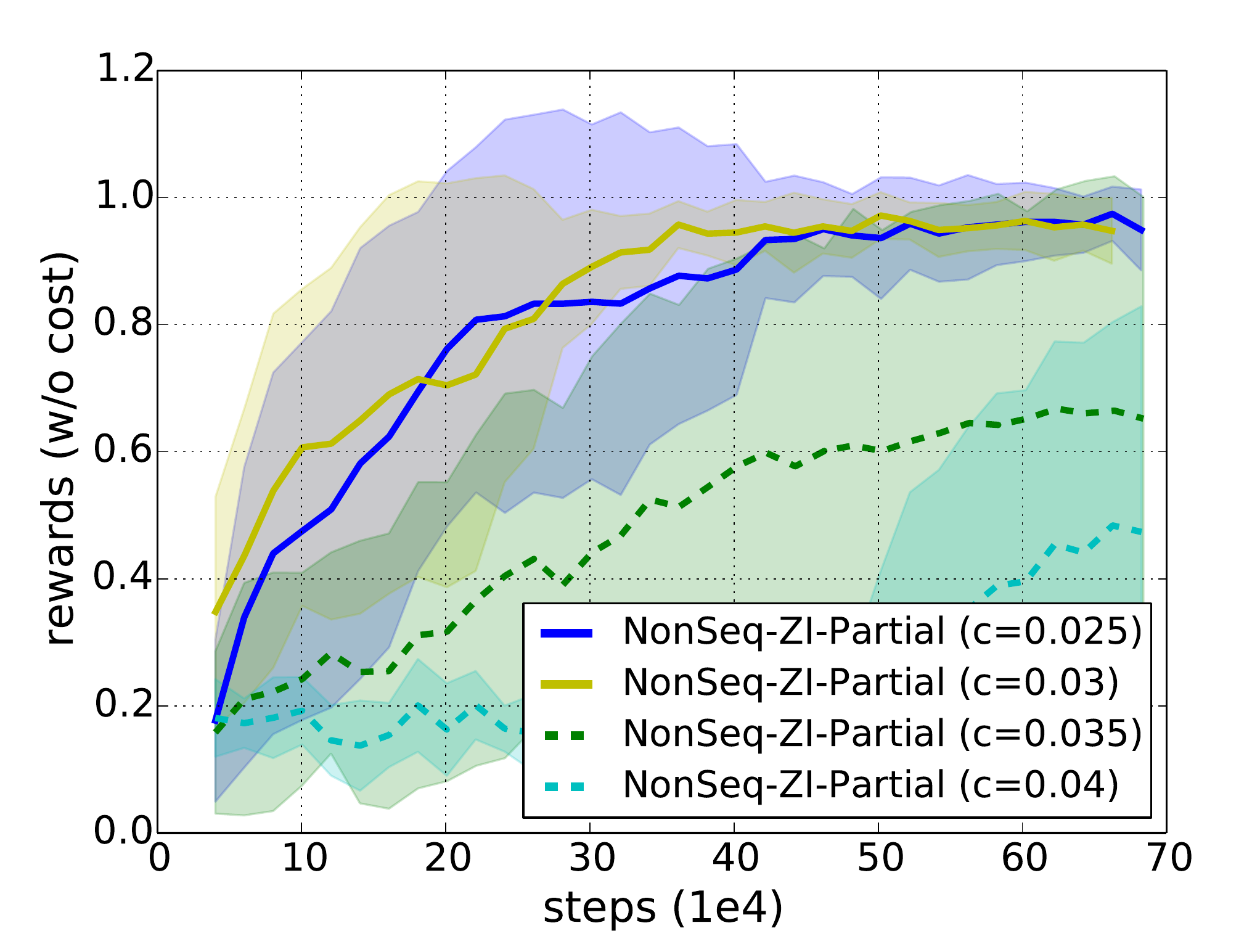} \hspace{4mm}
\includegraphics[width=0.3\textwidth]{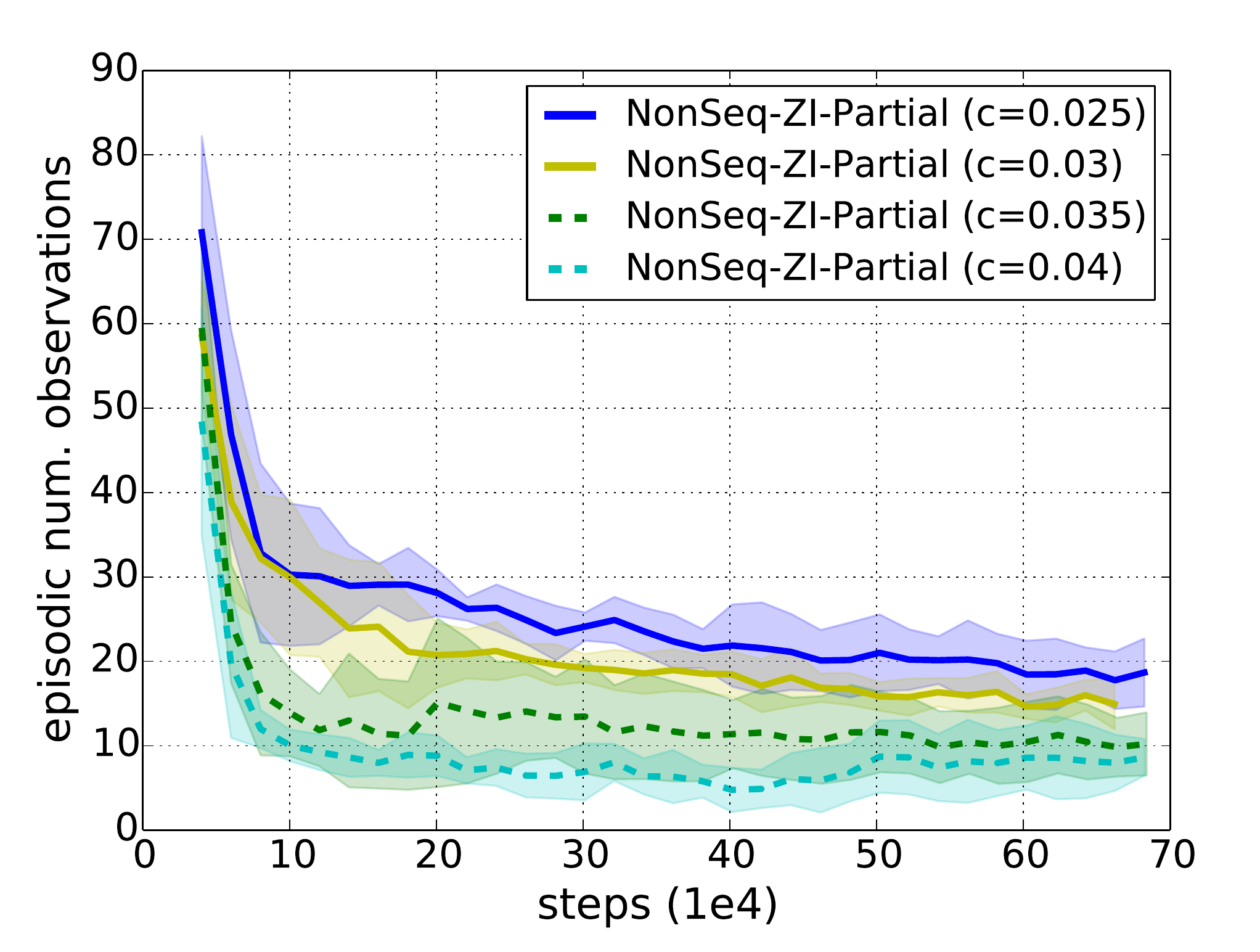}
}\\
\vskip -0.08in
\subfigure[Seq-PO-VAE (ours)]{\label{fig:ball_ours}
\includegraphics[width=0.3\textwidth]{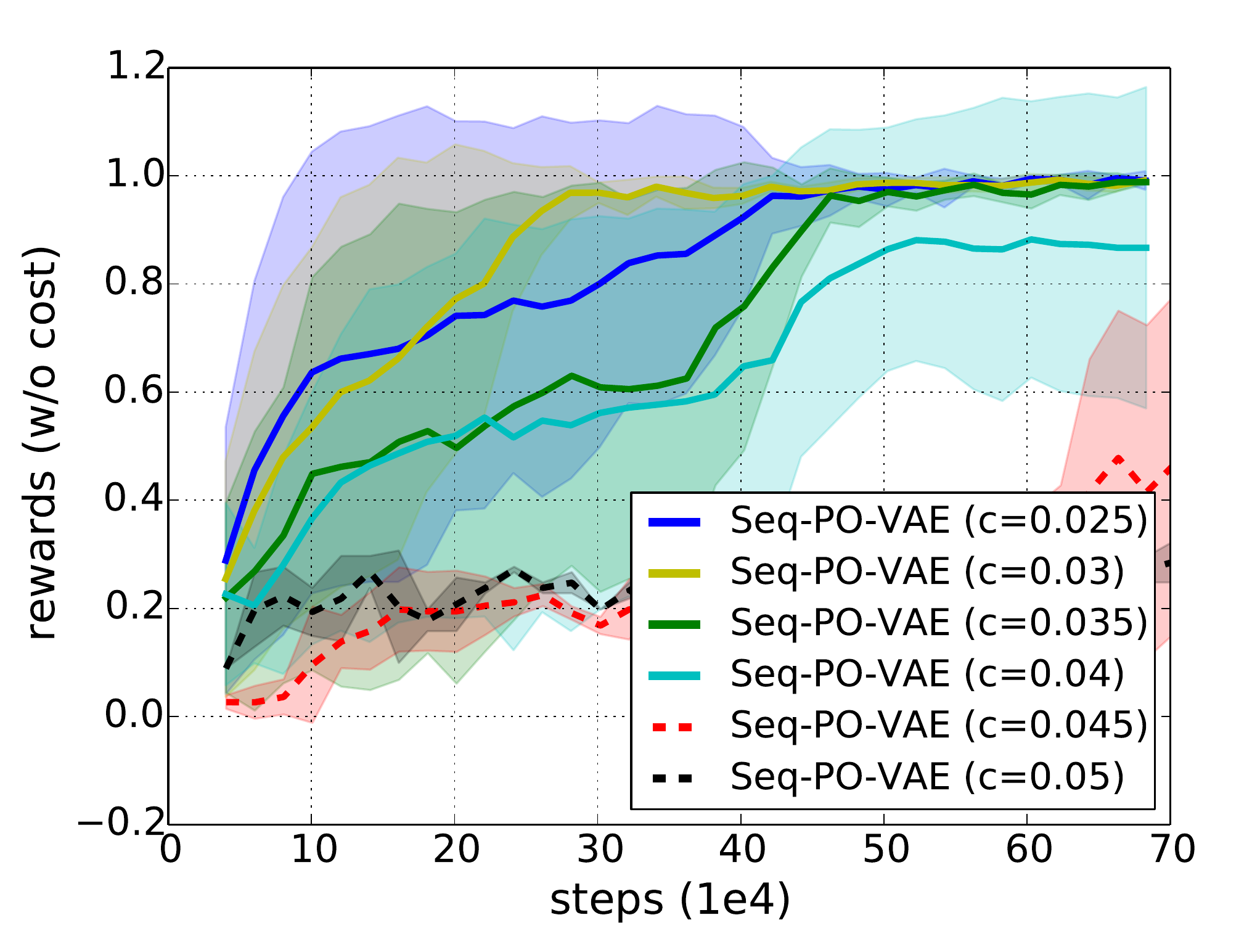} \hspace{4mm}
\includegraphics[width=0.3\textwidth]{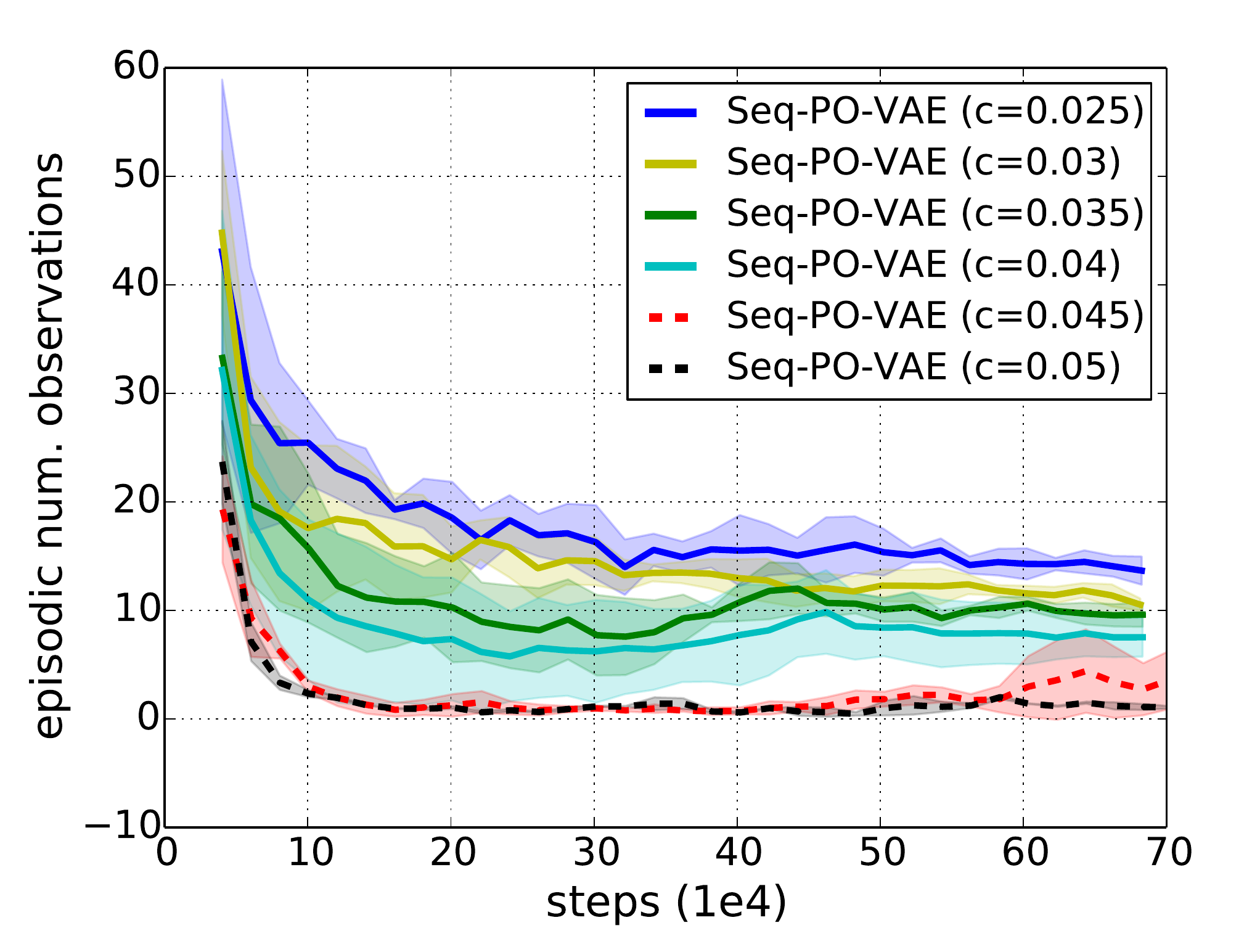}
}\\
\vskip -0.08in
\caption{Cost-performance trade-off investigation. Each row corresponds to the performance in terms of task reward (left) and number of acquisitions (per episode) obtained for a specific method (right), for a specific method (see the legend). Each curve is derived from 10 independent runs. We use dotted lines to indicate those instances for which the task learning does not always succeed. Thus, the best achievable number of observations should be referred to as the lowest curve among the \emph{solid} lines. Seq-PO-VAE consumes less than 10 observations to accomplish this task.  }
\label{fig:ball_cost}
\end{figure}

We perform a case study on investigating the cost-performance trade-off for each representation learning method and present the results in Figure~\ref{fig:ball_cost}. Apparently, as we increase the cost, the  exploration-exploitation task becomes more challenging and each compared method has its \emph{own upper limit of cost}, above which the model would fail to learn an effective task policy while acquiring minimum observations. First, we notice that the \emph{End-to-End} model takes a long time to progress in learning task skills (i.e., typically $> 1.5$m), while the VAE-based models can progress much faster. Among the VAE-based methods, we notice that our proposed method (Figure~\ref{fig:ball_ours}) can accomplish the task by acquiring as little as 8 observations whereas the baselines \emph{NonSeq-ZI (Full)} (Figure~\ref{fig:ball_nonseq_full}) and  \emph{NonSeq-ZI (partial)} (Figure~\ref{fig:ball_nonseq_partial}) achieve a standard of acquiring approximately 20 observations (refer to the lowest point among the \emph{solid} lines in the figure). Thus, we conclude that our proposed approach can significantly benefit the cost-sensitive policy training and leads to a policy which acquires fewer observations while achieving equal or better task performance. 
% \sebastian{Link text with subfigures?}

%% Sepsis Simulator
% %+++++++++++++++++++++++++++++++++++
\clearpage
\section{Sepsis Medical Simulator}
\label{appendix:sepsis}

\subsection{Task Specifications} 
For this task we employ a Sepsis simulator proposed in previous work~\cite{oberst2019counterfactual}. 
The task is to learn to apply three \emph{treatment} actions for Sepsis patients in intensive care units, i.e., $\mathcal{A}^c = \{$\emph{antibiotic}, \emph{ventilation}, \emph{vasopressors}$\}$. 
%Each treatment action is modeled as a Bernoulli variable. 
At each time step, the agent selects a subset of the \emph{treatment} actions to apply.
The state space consists of 8 features: 3 of them specify the current \emph{treatment} status; 4 of them specify the \emph{measurement} status in terms of \emph{heart rate}, \emph{sysBP rate}, \emph{percoxyg stage} and \emph{glucose level}; the remaining one is a categorical feature indicating the patent's antibiotic status.
The feature acquisition actively selects a subset among the \emph{measurement} features for observation, i.e., $\mathcal{A}^f = \{$\emph{heart rate}, \emph{sysBP rate}, \emph{percoxyg state}, \emph{glucose level}$\}$. The objective for learning an active feature acquisition strategy is to help the decision making system to reduce \emph{measurement} cost during its execution.

\subsection{Implementation Details}
For all the compared methods, we adopt \emph{Zero-Imputing}~\cite{nazabal2018handling} to fill in missing features. In particular, a fixed value of -10 which is outside the range of feature values is used to impute missing values.  

\paragraph{End-to-End} The end-to-end model first processes the imputed state by 3 \emph{fully connected} layers of size 32, 64 and 32, respectively. Each \emph{fully connected} layer is followed by a \emph{ReLU} activation function. 

\paragraph{NonSeq-ZI} The VAE model first processes the imputed state by 2 \emph{fully connected} layers with size 32 and 64, with the first \emph{fully connected} layer being followed by \emph{ReLU} activation functions. Then the output is fed into two independent \emph{fully connected} layers of size 10 for each, to generate the mean and variance for the Gaussian distribution. To decode the state, the latent code is first processed by a \emph{fully connected} layer of size 64, then fed into three \emph{fully connected} layers of size 64, 32, and 8. The intermediate \emph{fully connected} layers employ \emph{ReLU} activation functions. Also, we adopt two variants for \emph{NonSeq-ZI}, trained under either \emph{full} loss or \emph{partial} loss. The details of the hyperparameter settings used for training are presented in Table~\ref{table:sepsis_vae_params}.

\paragraph{Seq-PO-VAE (ours)} At each time step, the inputs for state and action are first processed by their corresponding projection layers. The projection layers for the state consists of 3 \emph{fully connected} layers of size 32, 16 and 10, where the intermediate \emph{fully connected} layers are followed by a \emph{ReLU}  activation function.
The projection layer for the action input is a \emph{fully connected} layer of size 10. Then the projected state feature $\f_x$ and action feature $\f_a$ are combined in the following manner: $\f_c = [\f_x,\, \f_a,\, \f_x*\f_a]$. $\f_c$ is passed to 2 \emph{fully connected} layers of size 64 and 32 to form the input to the \emph{LSTM} module. The output $\h_t$ of the \emph{LSTM} is fed to two independent \emph{fully connected} layers of size 10 to generate the mean and variance for the Gaussian distribution. The decoder for \emph{Seq-PO-VAE} has the identical architecture as that for \emph{NonSeq-ZI}. The details for training \emph{Seq-PO-VAE} are presented in Table~\ref{table:sepsis_vae_params}. 

\begin{table}[b!]
    \centering
    \caption{Hyperparameter settings for training VAE models on the \emph{Sepsis} task. }
    \label{table:sepsis_vae_params}
  \begin{tabular}{rcccc}
    \toprule
    & \multicolumn{4}{c}{Hyperparameter}\\\cmidrule{2-5}
     & $\beta$ (KL weight) & KL reduction & Loss reduction & learning rate  \\
    \midrule
    NonSeq-ZI (partial) & 0.01 & sum & sum & 1e-4 \\
    NonSeq-ZI (full) & 0.01 & sum & sum & 1e-4 \\
    Seq-PO-VAE (ours) & 0.01 & sum & sum & 1e-3 \\
    \bottomrule
  \end{tabular}
  
\end{table}

\paragraph{LSTM-A3C} The LSTM-A3C~\cite{mnih2016asynchronous} takes encoded state features derived from the corresponding representation model as its input. The encoded features are fed into an \emph{LSTM} with size 256. Then the $\h_t$ for the \emph{LSTM} is fed to three independent \emph{fully connected} layers, to predict the state value, feature acquisition policy and task policy. \emph{Normalized column} initialization is applied to all  \emph{fully connected} layers. The biases for the \emph{LSTM} and \emph{fully connected} layers are initialized as zero. 

\subsection{Data Collection} 
\label{app:sepsis:data_collect}
To train the VAEs, we prepare a training set that consists of 2000 trajectories. Half of the trajectories are derived from a random policy and the other half is derived from a policy learned \emph{End-to-End} with cost $0.0$. All the VAE models are evaluated on a test dataset that consists of identical size and data distribution as the the training dataset. We present the task treatment reward obtained by our data collection policy derived from the \emph{End-to-End} method and that obtained by our proposed method in Table~\ref{table:ball_task_obs}. Noticeably, by performing representation learning, our method could obtain much better treatment reward compared to the data collection policy. Therefore, it is essential to conduct representation learning to tackle the challenging AFA-POMDP problem.
\begin{table}[h!]
  \centering
  \caption{ Task performance for the data collection policy and our proposed method on \emph{Sepsis}.}
\label{table:ball_task_obs}
  \begin{tabular}{rcc}
    \toprule
    & \multicolumn{2}{c}{Model} \\\cmidrule{2-3}
     & End-to-End & Ours \\
    \midrule
    Treatment Reward & 0.35 & \textbf{0.45} \\
    \bottomrule
  \end{tabular}
\end{table}

\subsection{More Comparison Result under Different Values for Cost}

% We present the results for each method derived under the cost value of 0.01 in the main paper. We also show the performance curves for each independent run of our method in Figure~\ref{fig:sepsis_runs}. Overall, the results in Figure~\ref{fig:sepsis_runs} demonstrate that our method is quite stable and it converges to a better performance standard than the baseline methods in terms of \emph{discharge rate} and \emph{reward (w/o) cost} almost at each independent run.   

% \begin{figure*}[h]
% \centering
% % \vspace{3mm}
% \subfigure[Cost=0.1]{
% \includegraphics[width=0.32\textwidth]{latex/figures_appendix/appendix_sepsis_ours_discharge.pdf}
% \includegraphics[width=0.32\textwidth]{latex/figures_appendix/appendix_sepsis_ours_mortality.pdf}
% \includegraphics[width=0.32\textwidth]{latex/figures_appendix/appendix_sepsis_ours_reward.pdf}
% }
% \vskip -0.1in
% \caption{Result for 10 independent runs for our method.}
% \label{fig:sepsis_runs}
% \end{figure*}

\begin{figure*}[h!]
\centering
% \vskip -0.2in
\subfigure[Cost=0]{\label{fig:sepsis_cost_0}
\includegraphics[width=0.28\textwidth]{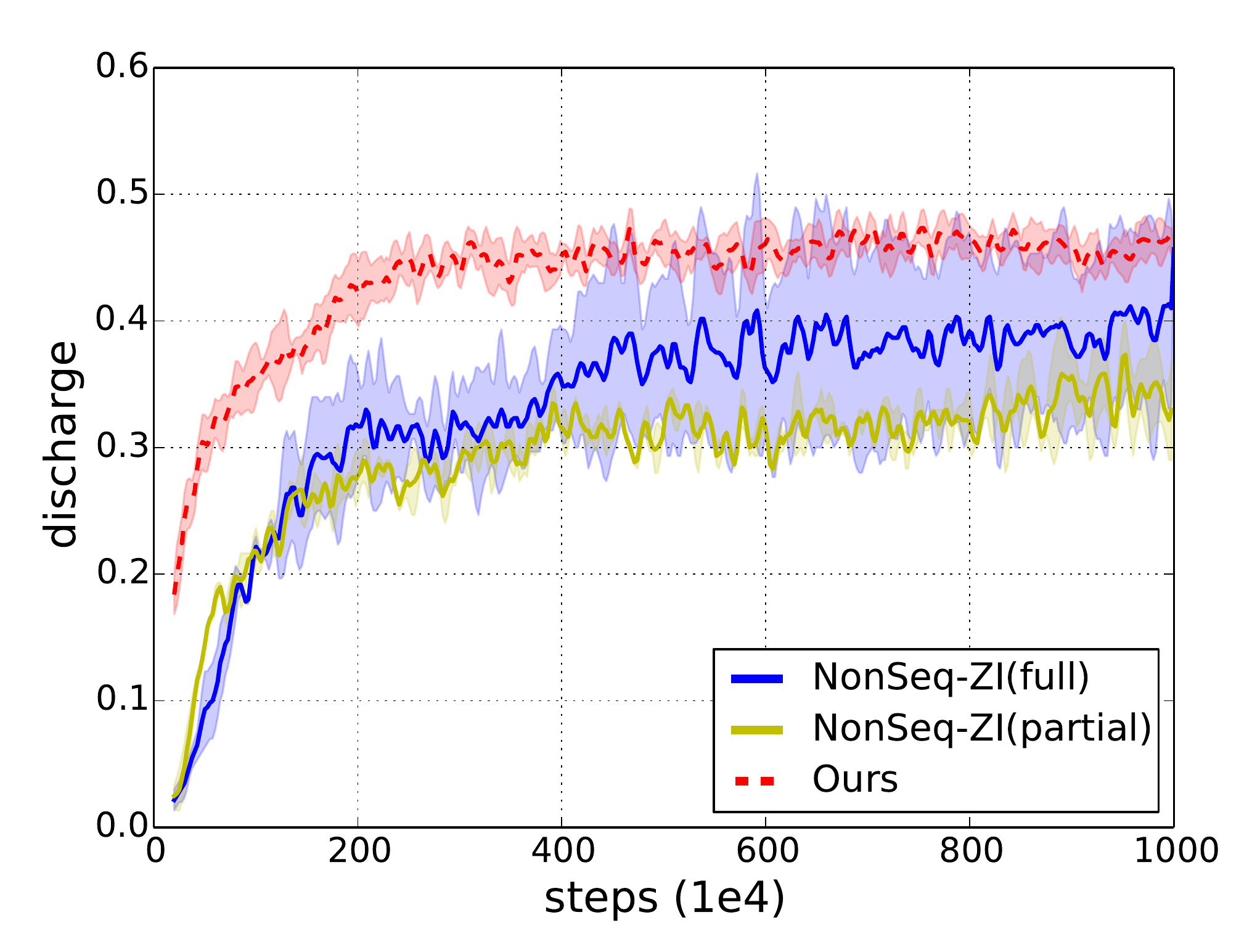}
\includegraphics[width=0.28\textwidth]{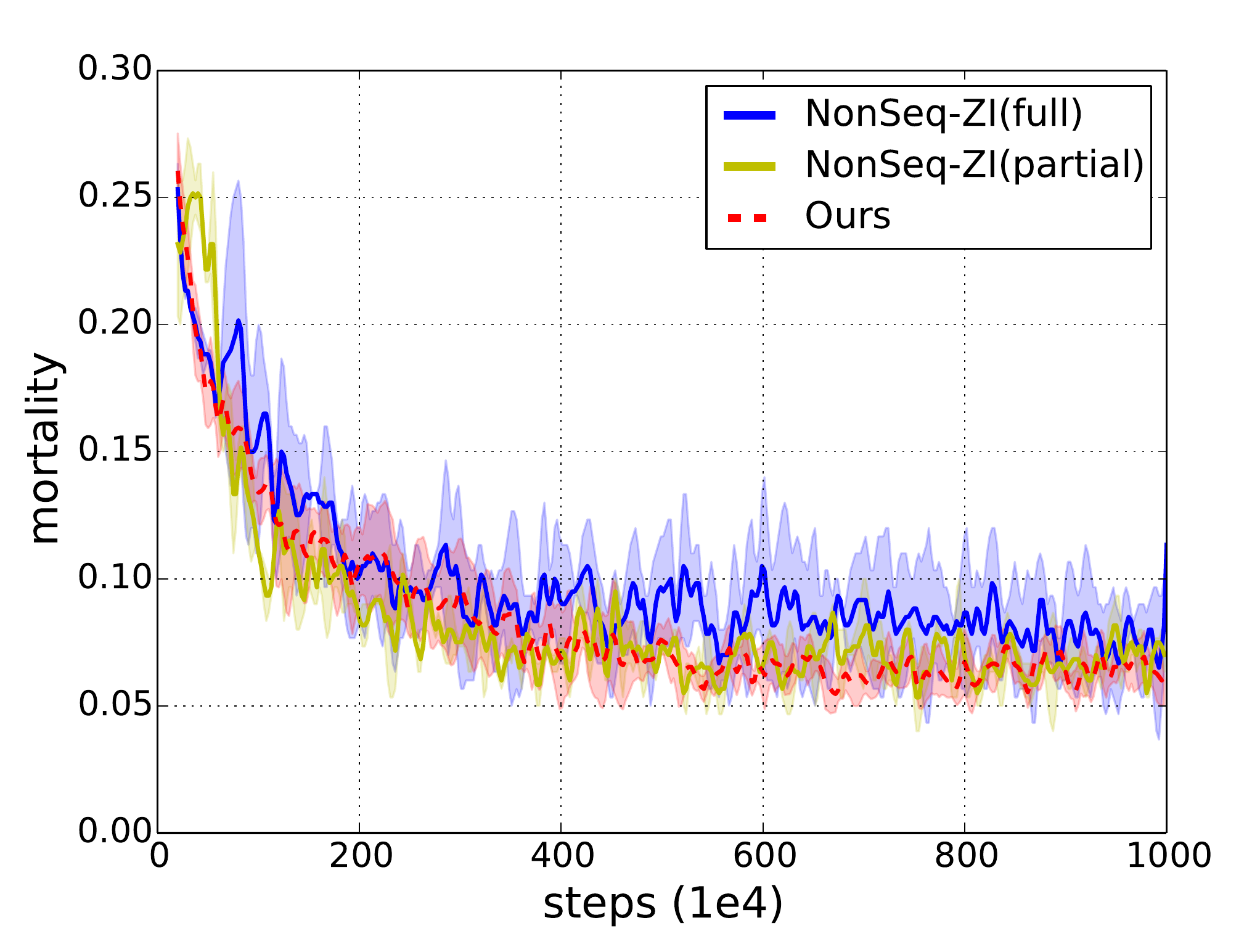}
\includegraphics[width=0.28\textwidth]{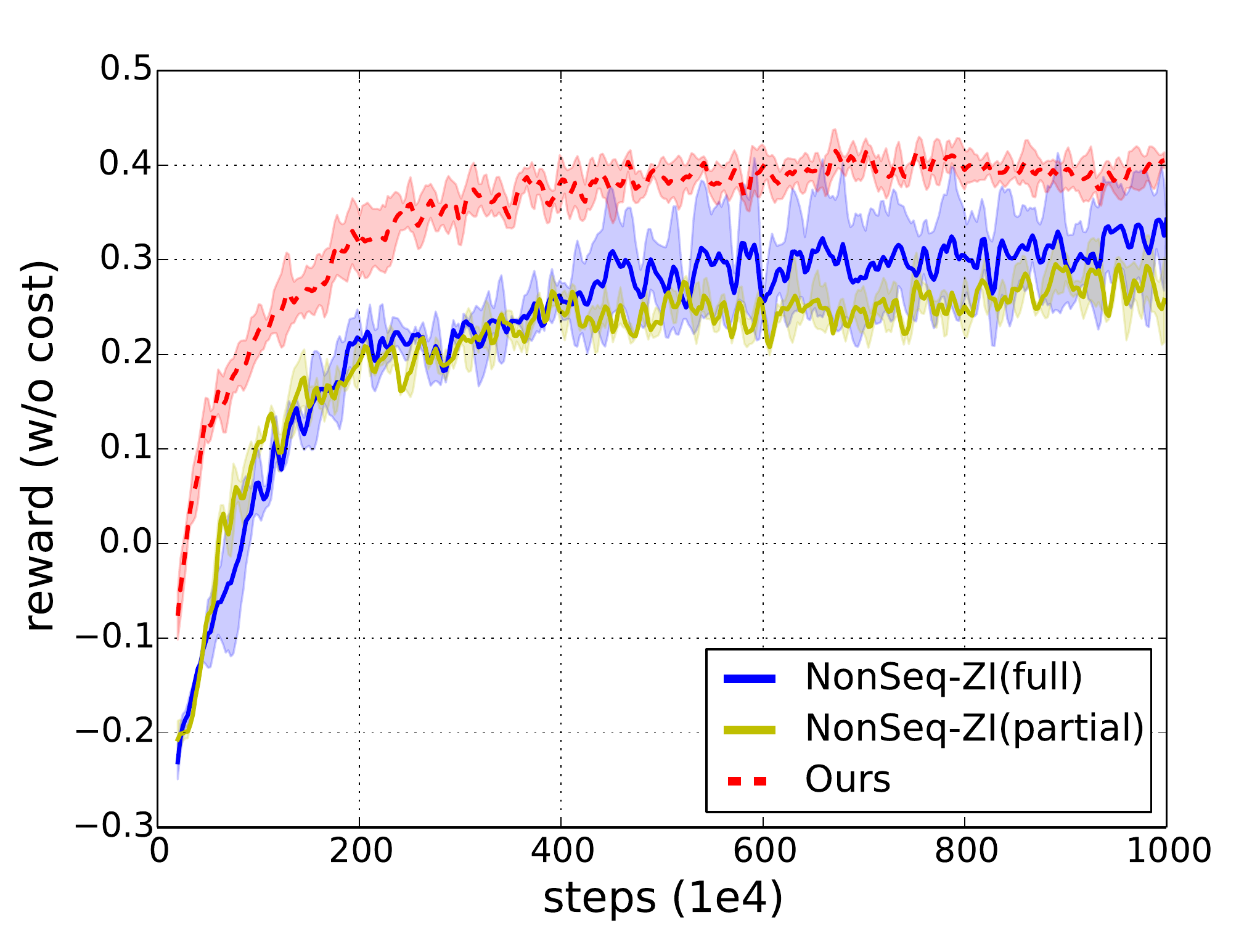}
}\\
\vskip -0.1in
\subfigure[Cost=0.025]{\label{fig:sepsis_cost_025}
\includegraphics[width=0.28\textwidth]{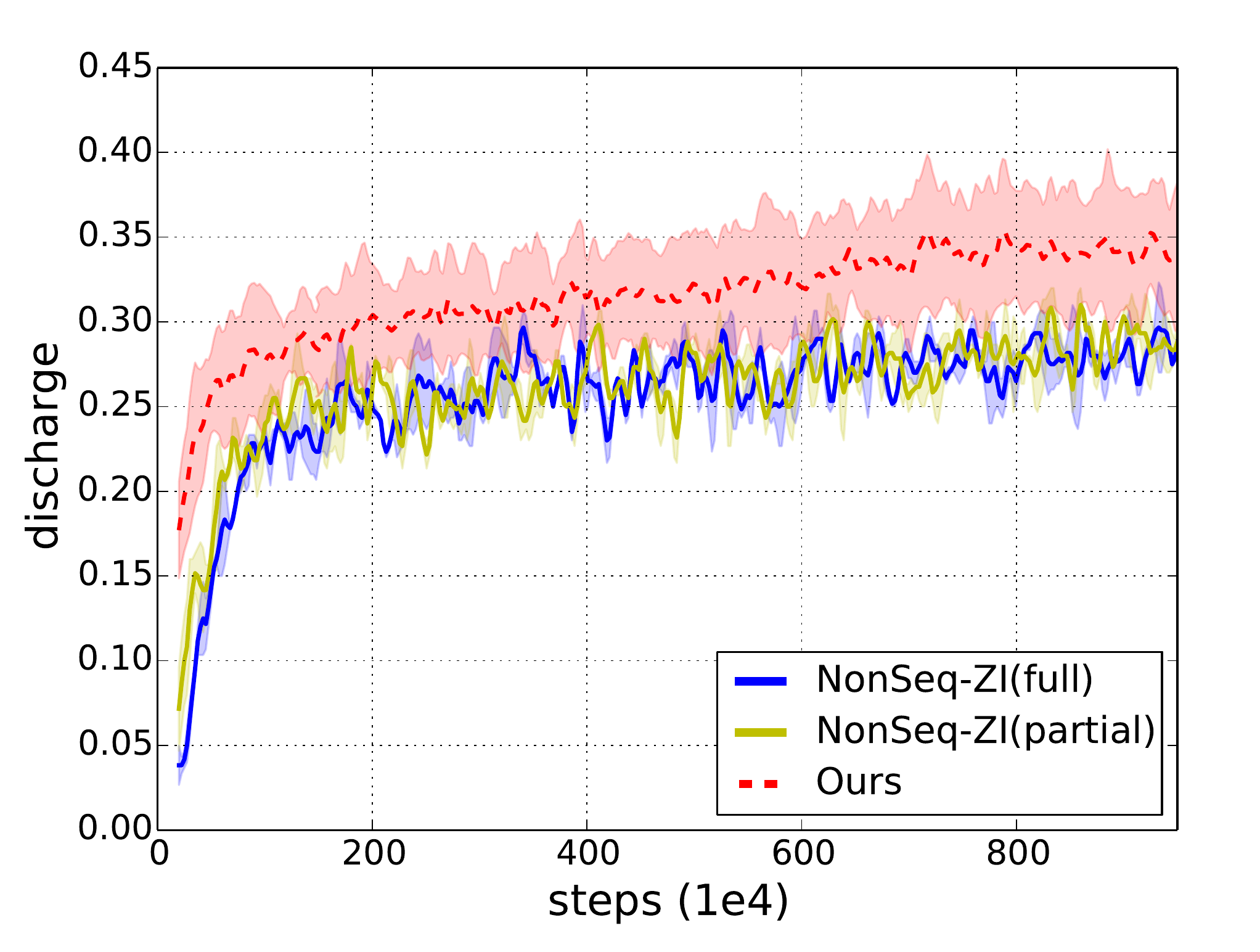}
\includegraphics[width=0.28\textwidth]{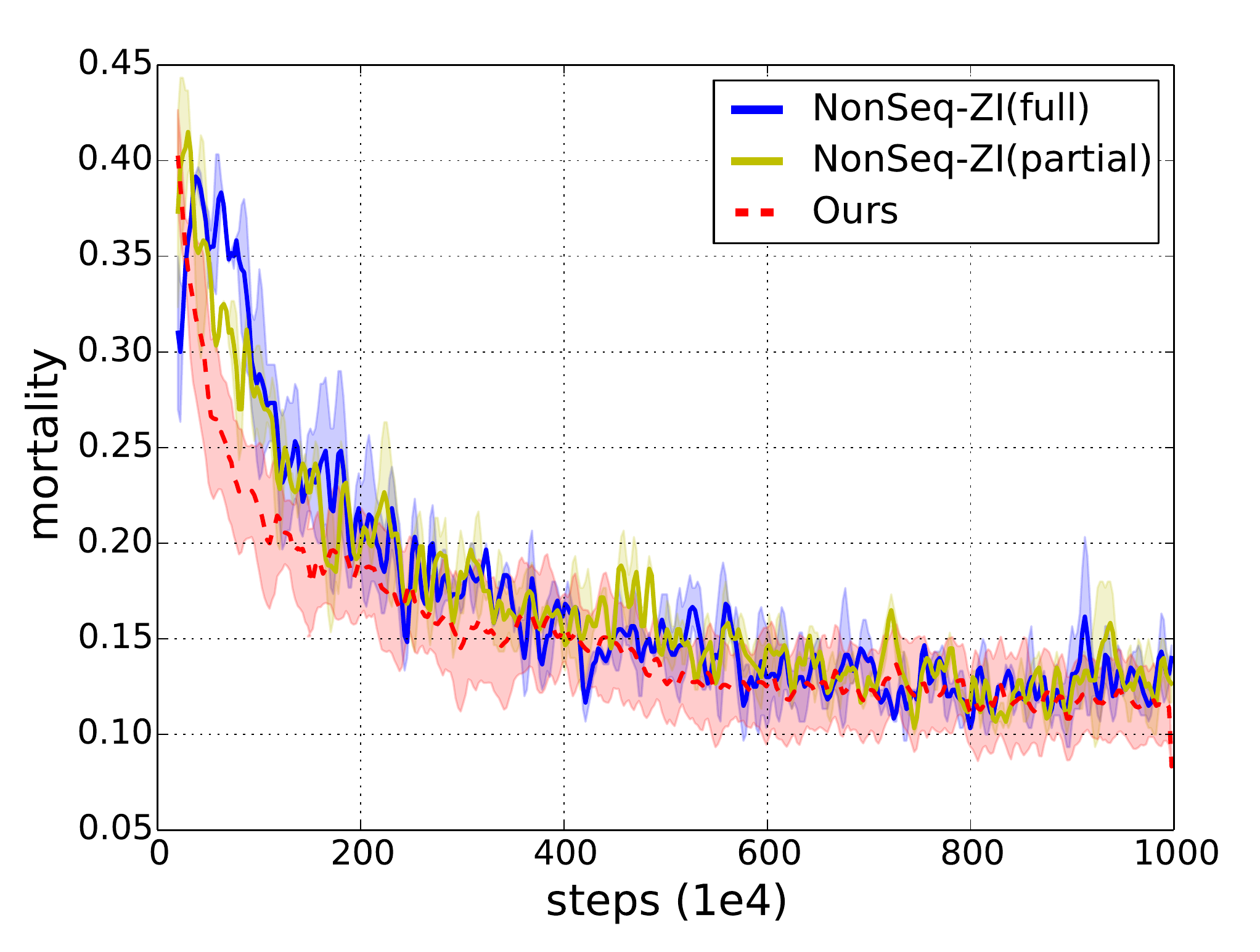}
\includegraphics[width=0.28\textwidth]{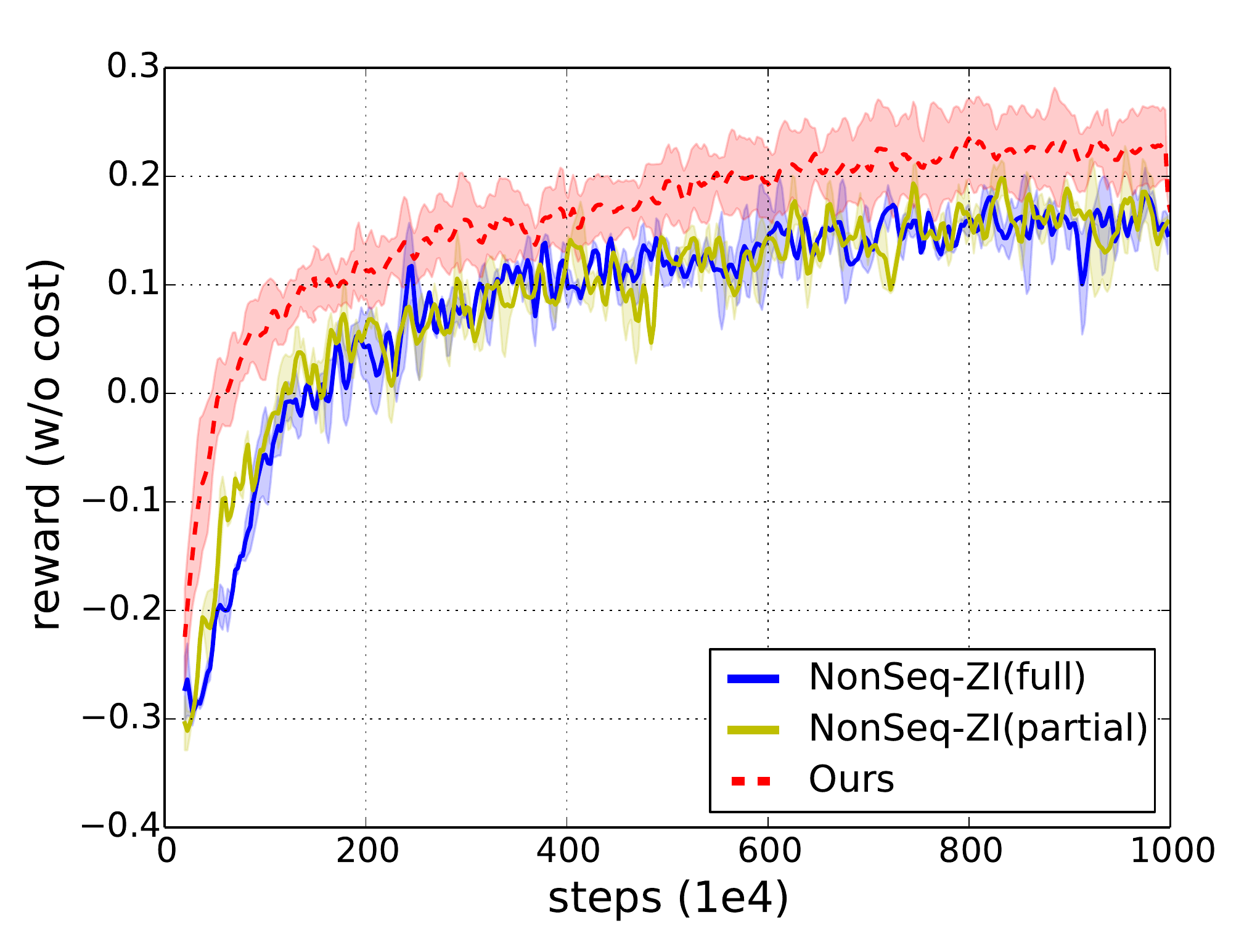}
}\\
\vskip -0.15in
\caption{Comparison result between our proposed method and the non-sequential VAE baseline models under different values for cost.  }
\end{figure*}

% \clearpage
We present the cost-performance trade-off on \emph{Sepsis} domain when running our method under different cost values in $\{$0, 0.025$\}$. The results are shown in Figure~\ref{fig:sepsis_cost_0} and Figure~\ref{fig:sepsis_cost_025}). By increasing the value of cost, we  obtain a feature acquisition policy that acquires substantially less features within each episode, with a sacrifice in task rewards. 

\begin{figure*}[h!]
\centering
\vskip -0.2in
\includegraphics[width=0.3\textwidth]{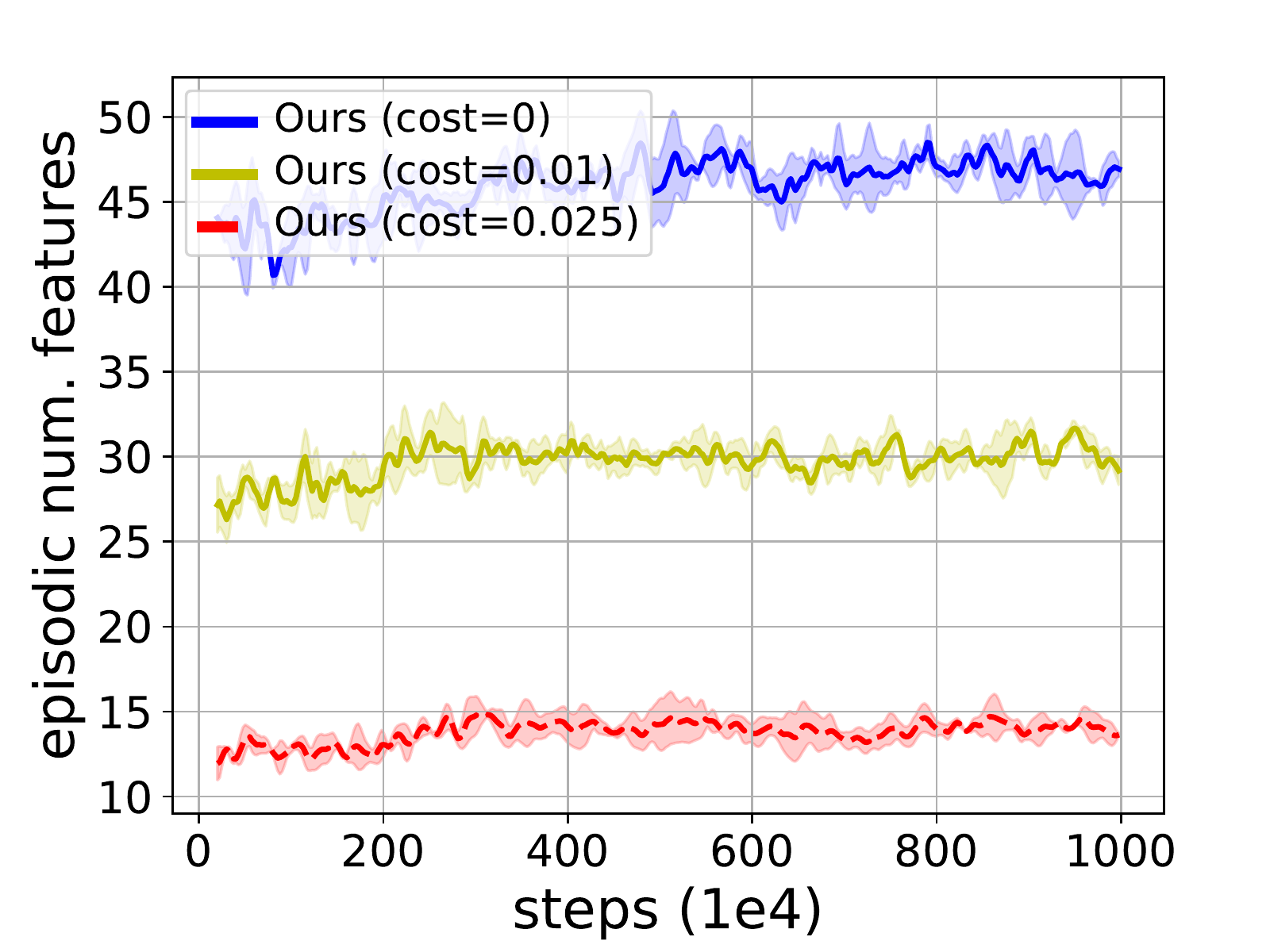}
\vskip -0.1in
\caption{Average num. observations acquired in each episode under cost values in \{0, 0.1, 0.025\}. }
\vskip -0.1in
\label{fig:sepsis_additional_cost}
\end{figure*}

% \begin{wrapfigure}{r}{0.6\columnwidth}
% \centering
% \vspace{-15pt}
% \includegraphics[width=0.3\textwidth]{latex/figures_appendix/Appendix_sepsis_num_obs.pdf}
% \vspace{-4mm}
% % \caption{Average number of observations acquired in each episode when training our proposed model under cost values 0 and 0.025. }
% \label{fig:sepsis_additional_cost}
% \vspace{-15pt}
% \end{wrapfigure} 
Furthermore, we present the episodic number of acquired features for our method in Figure~\ref{fig:sepsis_additional_cost}) when trained under different cost values. The results show that by increasing the cost, the number of feature acquisition substantially reduces. 

% \vskip -0.1in
\clearpage
\subsection{Illustrative Examples for Missing Feature Imputation in \emph{Sepsis}}
\begin{figure*}[h!]
\centering
\includegraphics[width=0.4\textwidth]{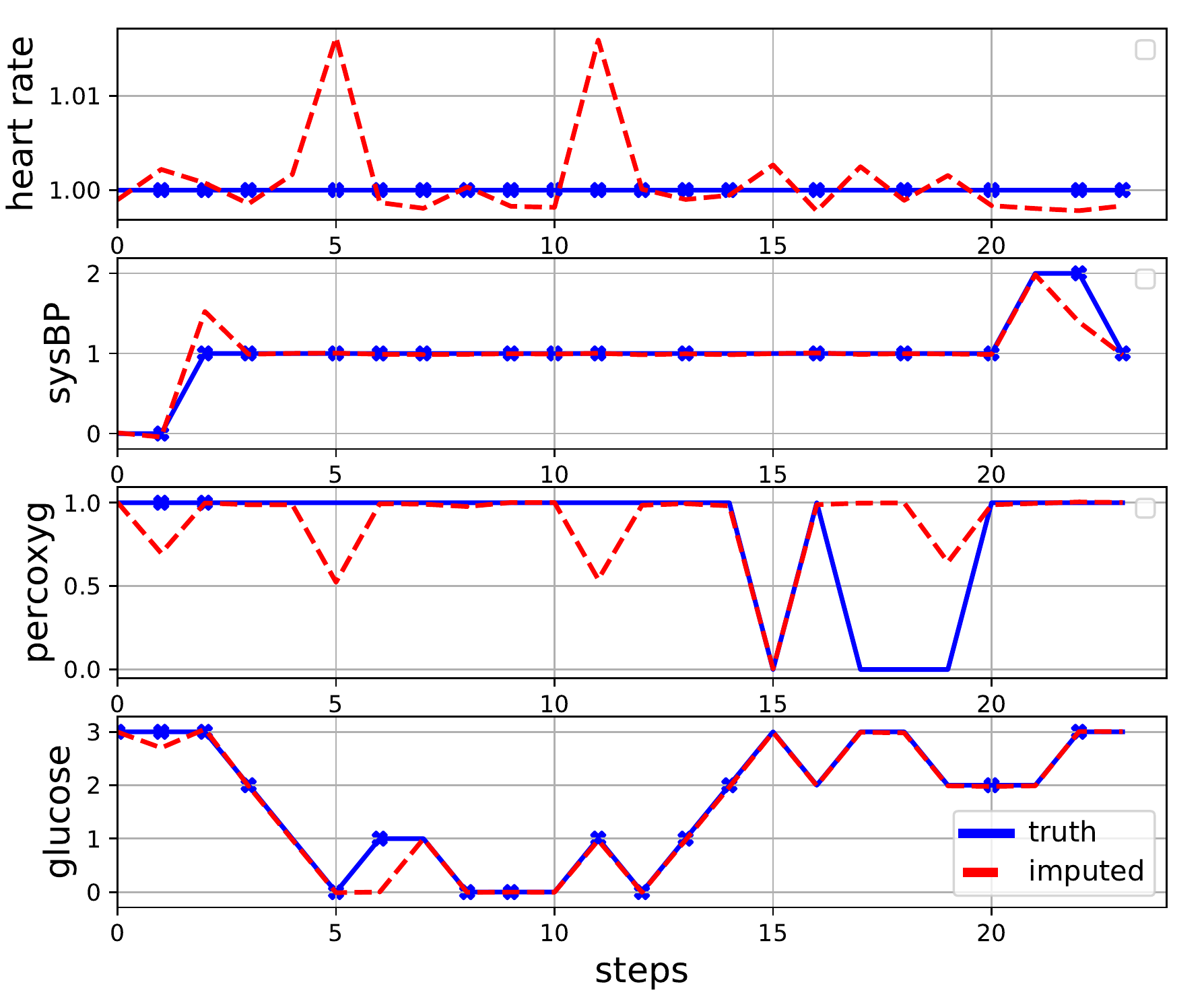} \hspace{0.3in}
\includegraphics[width=0.4\textwidth]{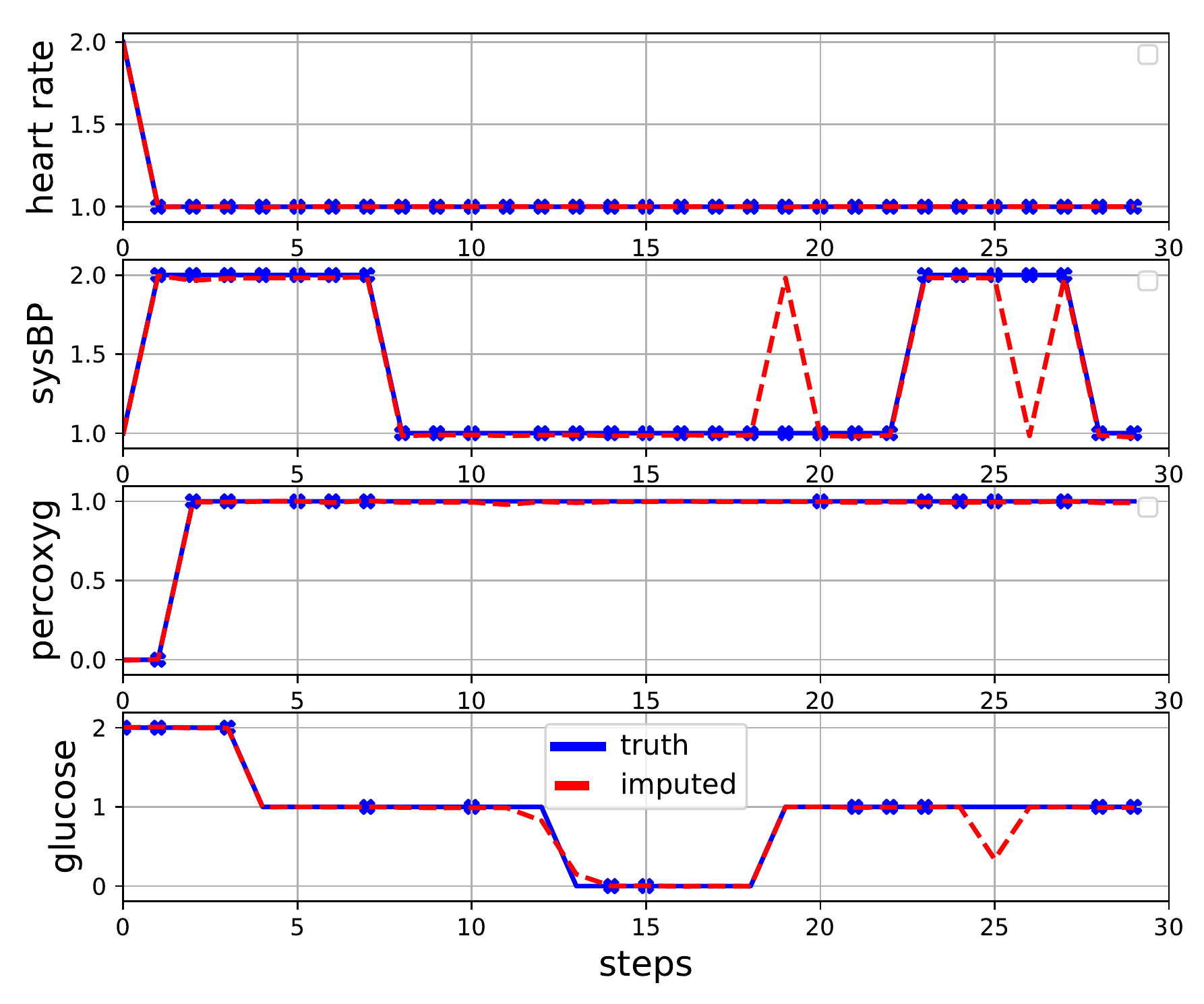}
\caption{Two example trajectories for illustrating how our method works on the \emph{Sepsis} medical domain. The acquisition policy is trained with a cost of 0. Each block corresponds to one trajectory and the four rows correspond to the four \emph{measurement} features being considered for active feature acquisition. Each dot indicates the employment of feature acquisition on the corresponding \emph{measurement} feature at the presented time point. In each trajectory, we demonstrate the ground-truth signal over time as well as the imputed signal over time predicted by our proposed  \emph{Seq-PO-VAE} model. By imputing the missing features via learning model dynamics, our proposed method could offer much more informative representation for the policy training compared to the non-sequential VAE baselines by giving reasonable imputation over the unobserved features.   }
\label{fig:sepsis_trajectory}
\end{figure*}

We present two illustrative examples in Figure~\ref{fig:sepsis_trajectory} to demonstrate how imputing missing features via learning model dynamics would help the decision making with partial observability in \emph{Sepsis} domain. The policy training process with partial observability can only access very limited information, due to the employment of active feature acquisition. Under such circumstances, imputing the missing features would offer much more abundant information to the decision making process. From the results shown in Figure~\ref{fig:sepsis_trajectory}, our model demonstrates considerable accuracy in imputing the missing features, even though it is extremely challenging to perform the missing feature imputation task given the distribution shift from the data collection policy and the online policy. The imputed missing information can be greatly beneficial for training the task policy and feature acquisition policy.

% \sebastian{Check which features are acquired. $\rightarrow$ Plot of signals over time by different policies?}
% \sebastian{Our notation is not consistent, we should ideally use bold variable names for vectors.}

% \clearpage
% \subsection{Imputing Missing Features with \emph{Seq-PO-VAE}}

%% Learning from Partial Observations
% \input{appendix_partial_obs.tex}
%+++++++++++++++++++++++++++++++++++
\section{Case Study: Investigating the Data Observability for Representation Learning}
% \chaptermark{A Case Study on Partial Data Observability}

In our proposed method, we assumed that the model has access to the fully observed data at the representation learning stage, so that the VAE can be trained to impute the missing features with the supervision of the fully observed data (following Equation (5) in the paper). In this section, we present a case study to demonstrate that such assumption does not necessarily need to hold and that our method can work with partially observed training data as well. 
To this end, we create two adapted baselines from our proposed method, where the representation learning models (i.e., Seq-PO-VAE) for the baselines are trained under partial observation, i.e., only 50\%/90\% of the features are accessible when training the Seq-PO-VAE model where the features to observe are randomly selected. We denote such adapted baselines as \emph{Seq-PO-VAE (50\%)} and \emph{Seq-PO-VAE (90\%)}, respectively.

\begin{table}[htbp]
\centering
\caption{Missing feature imputing loss evaluated on \emph{Bouncing Ball}$^+$ and \emph{Sepsis} domains.  }
\label{table:Observability}
% \begin{sc}
\scalebox{1.0}{
\begin{tabular}{lccr}
\toprule
\multirow{2}{*} {\textbf{VAE model} }    & \textbf{Bouncing Ball}$^+$ & \textbf{Sepsis} \\  & (NLL) & (MSE)  \\
\midrule
\multirow{2}{*} {NonSeq-ZI (partial)}   & 0.6504  & 0.8441   \\ & ($\pm$ 0.1391)  & ($\pm$0.0586) \\
% \midrule
\multirow{2}{*} {NonSeq-ZI (full)}    & 0.0722 & 0.4839 \\  & ($\pm$ 0.0004) & ($\pm$ 0.0012)  \\
% NonSeq-PN (full)   &  & \\
% NonSeq-PN (partial)   &  & \\
\midrule
\multirow{2}{*} {Seq-PO-VAE (50\%)}   & 0.0375 & 0.2892 \\ & ($\pm$ 0.0010) & ($\pm$ 0.0097 ) \\
\multirow{2}{*} {Seq-PO-VAE (90\%)}   & 0.0381  & 0.2450  \\ & ($\pm$ 0.0015 ) & ($\pm$ 0.0096 ) \\
\multirow{2}{*} {Seq-PO-VAE (full)}   & \textbf{0.0324} &  \textbf{0.1832} \\ & ($\pm$ 0.0082) & ($\pm$0.0158) \\
\bottomrule
\end{tabular}
}
% \end{sc}
\end{table} 

We present the missing feature imputing performance for the VAE models evaluated on the two task domains in Table~\ref{table:Observability}. From the results, we notice that with reduced observability, the missing feature imputing performance for \emph{Seq-PO-VAE (50\%/90\%)} degrades to fall below \emph{Seq-PO-VAE (full)}, which is as expected. However, the adapted baselines with partial observability can still benefit from our proposed sequential modeling with dynamics learning a lot. As a result, \emph{Seq-PO-VAE (50\%/90\%)} can outperform the non-sequential baselines \emph{NonSeq-ZI (partial/full)} on both missing feature imputing tasks with substantial performance margins. Note that the model \emph{NonSeq-ZI (full)} still employs full observation over the dataset during its training, but its missing feature imputing performance is substantially inferior as compared to \emph{Seq-PO-VAE (50\%)}. Overall, the above results demonstrate that our proposed representation learning method can derive meaningful representation with considerable efficiency in imputing missing features even when the model is trained under partial observation.

\begin{figure}[htbp]
	\centering
	\includegraphics[width=.3\columnwidth]{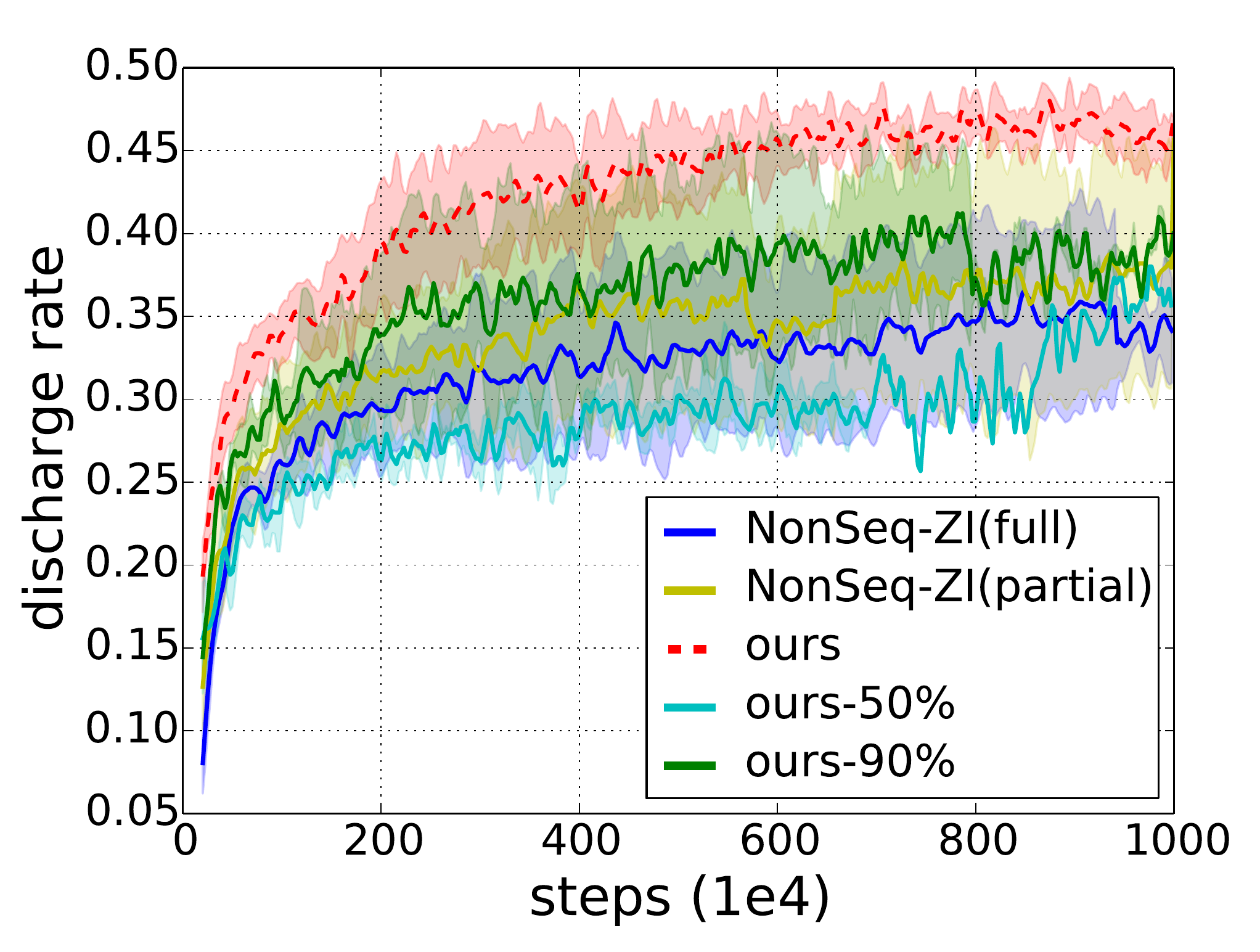} %\hspace{2mm}
	\includegraphics[width=.3\columnwidth]{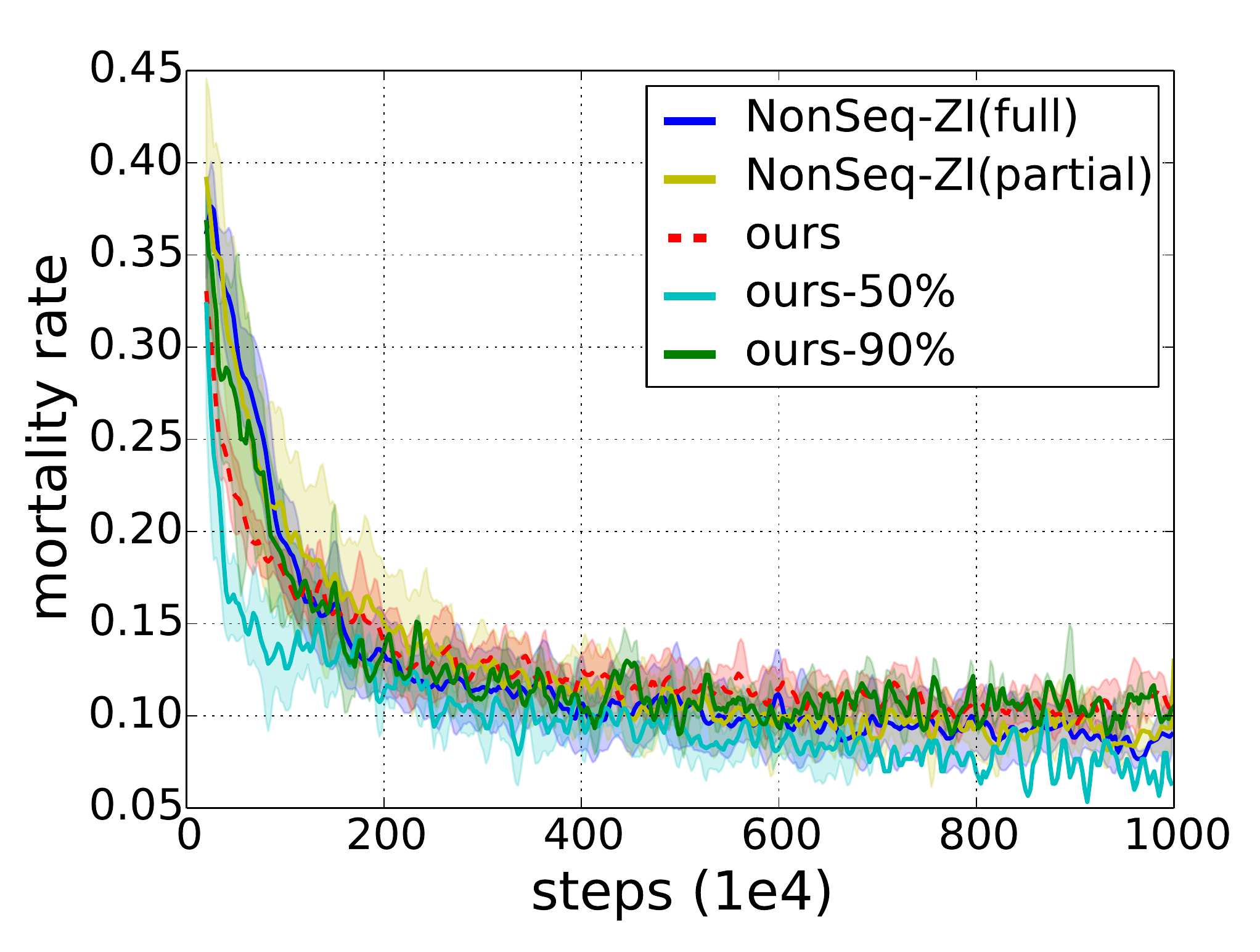}
	\includegraphics[width=.3\columnwidth]{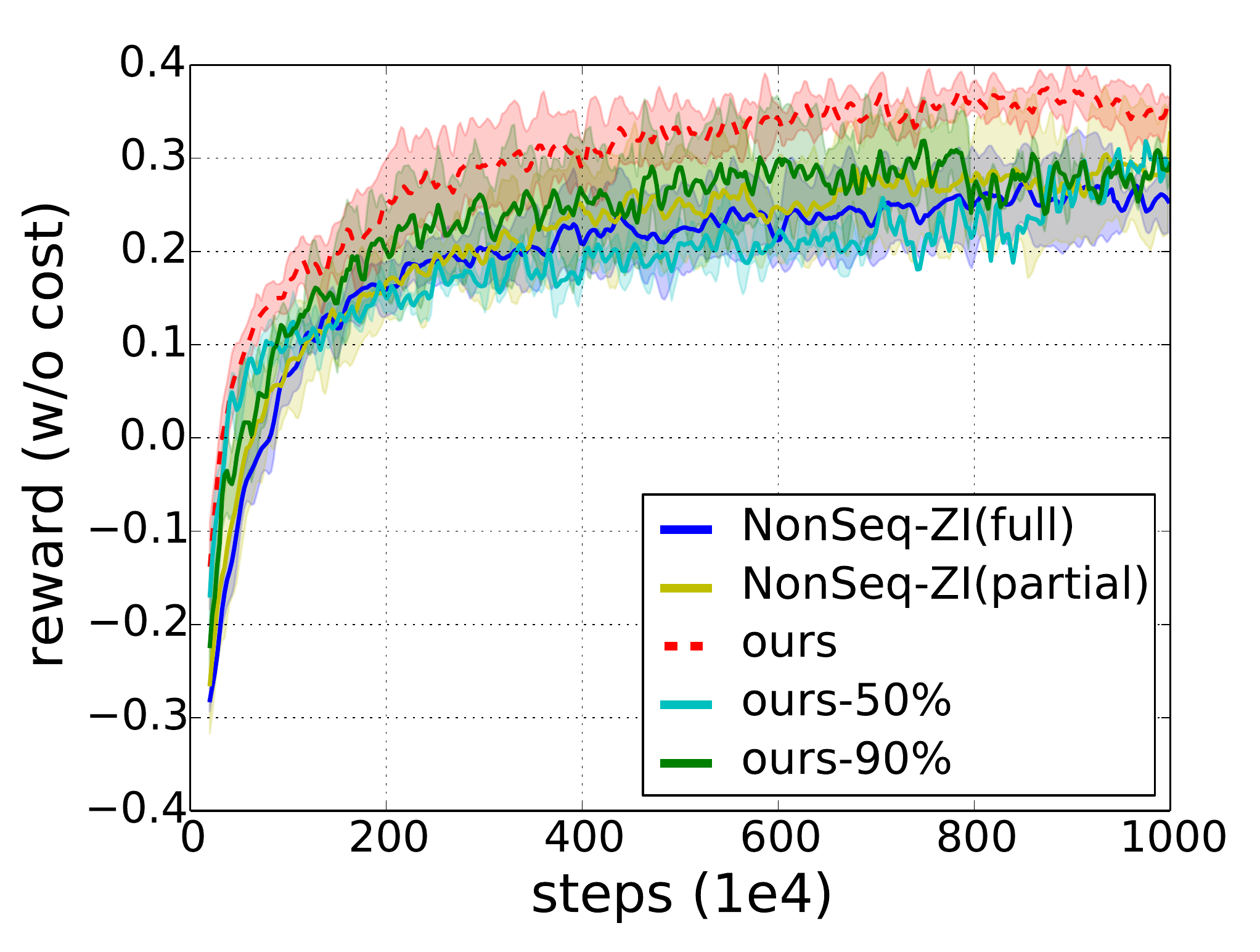}
	\caption{Performance curves in terms of \emph{discharge rate}, \emph{mortality rate} and \emph{reward (w/o cost)} on \emph{Sepsis} domain, evaluated with a cost value of 0.01. 
	}
	\label{fig:sepsis_po_perf}
\end{figure}

Furthermore, we demonstrate the policy training performance for the  \emph{Seq-PO-VAE (50\%/90\%)} baselines evaluated on the \emph{Sepsis} domain. The results are shown in Figure~\ref{fig:sepsis_po_perf}.  As expected, the performance of  \emph{Seq-PO-VAE} trained with partial observation degrades from that trained with full observation. The reason is due to that the task of imputing the missing features via learning system dynamics could be extremely challenging when only partial features are presented during training. However, when the level of observability is high, the model can still lead to promising performance that outperforms the non-sequential VAE baselines. Overall, the results reveal that our proposed method works best with full observability, but it is promising to work with partial observability when the level of observability is relatively high. Adapting our proposed method to tackle challenging AFA-POMDP domains with restricted level of observability to data is subject to future work, and our approach will benefit from any advances in representation learning from partially observed data.

\end{appendices}

\end{document}